\documentclass{article}

\usepackage{microtype}
\usepackage{graphicx}
\usepackage{subfigure}
\usepackage{booktabs} 
\usepackage{url}

\usepackage{amsmath,amsfonts,bm}

\def\eqref#1{equation~\ref{#1}}

\def\1{\bm{1}}

\DeclareMathAlphabet{\mathsfit}{\encodingdefault}{\sfdefault}{m}{sl}
\SetMathAlphabet{\mathsfit}{bold}{\encodingdefault}{\sfdefault}{bx}{n}

\newcommand{\E}{\mathbb{E}}

\usepackage{amsmath}
\usepackage{amssymb}
\usepackage{mathtools}
\usepackage{amsthm}
\usepackage{wrapfig}
\usepackage{caption}
\usepackage{subcaption}
\usepackage{enumitem}
\usepackage{graphicx}
\usepackage{xspace}
\usepackage{algorithmic}
\usepackage[ruled,vlined]{algorithm2e}
\usepackage{multicol}
\usepackage{multirow}
\usepackage{booktabs}
\usepackage{mathtools}
\usepackage{tcolorbox}
\usepackage{enumitem}
\def\method{VideoAgent\xspace}
\usepackage[compact]{titlesec} 
\usepackage[pagebackref=true,colorlinks,citecolor=brown]{hyperref}

\definecolor{verylightgrey}{rgb}{0.95,0.95,0.95}
\definecolor{darkgrey}{rgb}{0.25,0.25,0.25}

\newtcolorbox{promptbox}[2][]{
  colback=verylightgrey,
  colframe=black,
  coltitle=white,
  colbacktitle=darkgrey,
  boxrule=0.5pt,
  arc=5mm,
  outer arc=5mm,
  leftrule=1pt,
  rightrule=1pt,
  toprule=1pt,
  bottomrule=1pt,
  left=10pt,
  right=10pt,
  top=10pt,
  bottom=10pt,
  boxsep=0pt,
  title={\centering\strut#2\strut},
  fonttitle=\bfseries\large,
  #1
}

\usepackage{colortbl}

\usepackage{hyperref}

\usepackage[accepted]{icml2025}

\usepackage{amsmath}
\usepackage{amssymb}
\usepackage{mathtools}
\usepackage{amsthm}

\usepackage[capitalize,noabbrev]{cleveref}

\theoremstyle{plain}

\theoremstyle{definition}

\theoremstyle{remark}

\usepackage[textsize=tiny]{todonotes}

\usepackage{fnpct}
\icmltitlerunning{VideoAgent: Self-Improving Video Generation for Embodied Planning}

\begin{document}

\twocolumn[
\icmltitle{VideoAgent: Self-Improving Video Generation for Embodied Planning}



\icmlsetsymbol{equal}{*}

\begin{icmlauthorlist}
\icmlauthor{Achint Soni}{yyy,equal}
\icmlauthor{Sreyas Venkataraman}{zzz,equal}
\icmlauthor{Abhranil Chandra}{yyy,equal}
\icmlauthor{Sebastian Fischmeister}{yyy}
\icmlauthor{Percy Liang}{bbb}
\icmlauthor{Bo Dai}{aaa,comp}
\icmlauthor{Sherry Yang}{bbb,comp,sch}
\end{icmlauthorlist}

\icmlaffiliation{yyy}{University of Waterloo}
\icmlaffiliation{zzz}{ IIT Kharagpur}
\icmlaffiliation{aaa}{Georgia Institute of Technology}
\icmlaffiliation{bbb}{Stanford University}
\icmlaffiliation{comp}{Google Deepmind}
\icmlaffiliation{sch}{New York University}

\icmlcorrespondingauthor{Achint Soni}{achint.s046@gmail.com}
\icmlcorrespondingauthor{Sherry Yang}{sherryy@google.com}

\icmlkeywords{Machine Learning, ICML}

\vskip 0.3in
]



\printAffiliationsAndNotice{\icmlEqualContribution} 

\begin{abstract}
Video generation has been used to generate visual plans for controlling robotic systems. Given an image observation and a language instruction, previous work has generated video plans which are then converted to robot controls to be executed. However, a major bottleneck in leveraging video generation for control lies in the quality of the generated videos, which often suffer from hallucinatory content and unrealistic physics, resulting in low task success when control actions are extracted from the generated videos. While scaling up dataset and model size provides a partial solution, integrating external feedback is both natural and essential for grounding video generation in the real world. With this observation, we propose \method for self-improving generated video plans based on external feedback. Instead of directly executing the generated video plan, \method first refines the generated video plans using a novel procedure which we call \emph{self-conditioning consistency}, allowing inference-time compute to be turned into better generated video plans. As the refined video plan is being executed, \method can collect additional data from the environment to further improve video plan generation. Experiments in simulated robotic manipulation from MetaWorld and iTHOR show that \method drastically reduces hallucination, thereby boosting success rate of downstream manipulation tasks. We further illustrate that \method can effectively refine real-robot videos, providing an early indicator that robots can be an effective tool in grounding video generation in the physical world. Video demos and code can be found at \url{https://video-as-agent.github.io}. 
\end{abstract}

\setlength{\abovedisplayskip}{1pt}
\setlength{\abovedisplayshortskip}{1pt}
\setlength{\belowdisplayskip}{1pt}
\setlength{\belowdisplayshortskip}{1pt}
\setlength{\jot}{1pt}

\setlength{\parskip}{0.28em}
\titlespacing\section{0pt}{5pt plus 2pt minus 2pt}{2pt plus 2pt minus 2pt}
\titlespacing\subsection{0pt}{5pt plus 2pt minus 2pt}{2pt plus 2pt minus 2pt}
\makeatletter
\renewcommand{\paragraph}{%
  \@startsection{paragraph}{4}%
  {\z@}{0.05ex \@plus .05ex \@minus .05ex}{-1em}%
  {\normalfont\normalsize\bfseries}%
}

\section{Introduction}
\label{introduction}

Large text-to-video models pretrained on internet-scale data have broad applications such as generating creative video content~\cite{ho2022imagen,hong2022cogvideo,singer2022make} and creating novel games~\cite{bruce2024genie}, animations~\cite{wang2019learning}, and movies~\cite{zhu2023moviefactory}. Furthermore, recent work show that video generation can serve as simulators of the real-world~\cite{yang2023learning,videoworldsimulators2024}, as well as policies with unified observation and action space~\cite{du2024learning,ko2023learning,du2023video}. These recent applications of text-to-video generation models holds great promise of internet-scale knowledge transfer (e.g., from generating human videos to generating robot videos), as well as paving the way to generalist agent (e.g., a single policy that can control multiple robots with different morphologies in different environments to perform diverse tasks). 

Nevertheless, text-to-video models have only had limited success in downstream applications in reality. For instance, in video generation as policy~\cite{du2024learning,ko2023learning}, when an observation image and a language instruction are given to a video generation model, generated videos often hallucinate (e.g., objects randomly appear or disappear) or violate physical laws (e.g., a robot hand going through an object)~\cite{yang2023learning,videoworldsimulators2024}. Such hallucinations and unrealistic physics have led to low task success rate when generated videos are converted to control actions through inverse dynamics models, goal conditioned policies, or other action extraction mechanisms~\cite{wen2023any,yang2024video,ajay2024compositional}. 

While scaling up dataset and model size can be effective in reducing hallucination in large language models (LLMs)~\cite{hoffmann2022training}, scaling is more difficult in video generation models. This is partially because language labels for videos are labor intensive to curate. Moreover, video generation has not converged to an architecture that is more favourable to scaling~\cite{yang2024video}. Scaling aside, being able to incorporate external feedback to improve generation is one of the other most important breakthrough in LLMs~\cite{ouyang2022training}. It is therefore natural to wonder what kind of feedback is available for text-to-video models, and how we can incorporate these feedback to further improve the quality of the generated videos.

\begin{figure*}[t]
    \centering
    \includegraphics[width=1\linewidth]{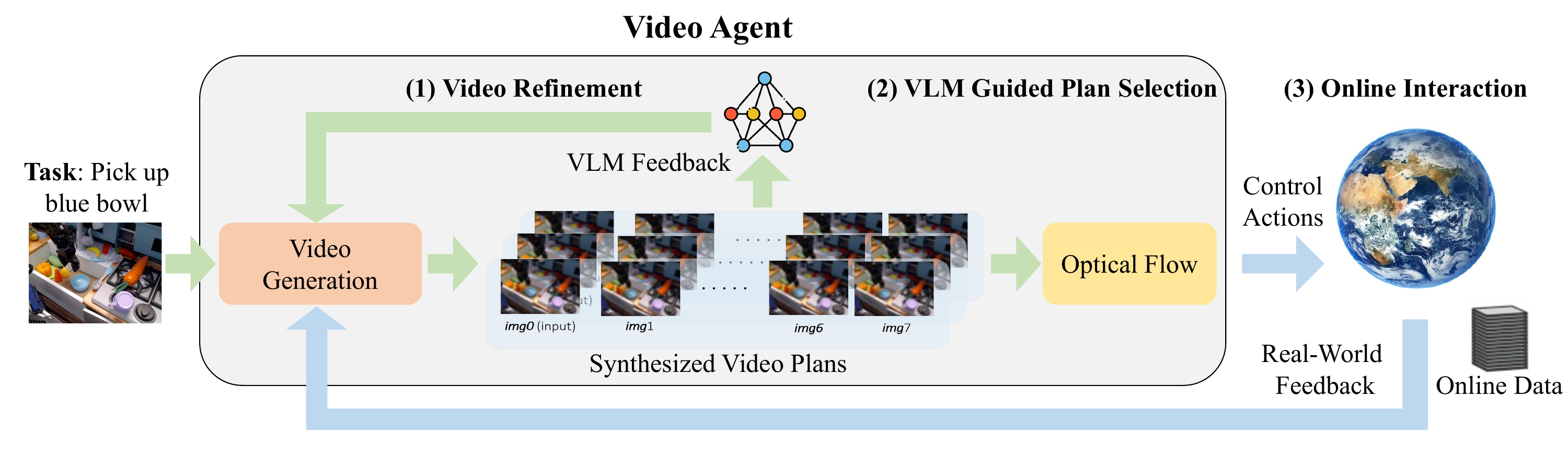}
    \caption{\textbf{The VideoAgent Framework.} \method first generates a video plan conditioned on an image observation and task description similar to \citet{du2023video}, and undergoes (1) iterative video refinement using feedback from a vision language model (VLM), (2) using the VLM to select the best refined video plan to convert to control actions through optical flow, and (3) executing the control actions in an environment and improving video generation using real-world feedback and additional data collected online.}
    \label{fig:framework}
\end{figure*}

To answer this question, we explore two types of feedback that are natural to acquire for video generation models, namely AI feedback from a vision-language model (VLM) and real-world execution feedback when generated videos are converted to motor controls. To utilize these feedback for self-improvement, we propose \method. Different from video generation 
as policy, which directly turns a generated video into control actions~\cite{du2023video,ko2023learning}, \method is trained to refine a generated video plan iteratively using feedback from a pretrained VLM. During inference, \method queries the VLM to select the best refined video plan, allowing inference-time compute to be turned into better generated video plans, followed by execution of the plan in the environment. During online execution, \method observes whether the task was successfully completed, and further improves the video generation model based on the execution feedback from the environment and additional data collected from the environment. The improvement to the generated video plan comes in three folds: First, we propose \emph{self-conditioning consistency} for video diffusion model inspired by consistency models~\cite{song2023consistency,heek2024multistep}, which enables low-quality samples from a video diffusion model to be further refined into high-quality samples. Second, VLM feedback combined with more inference-time compute leads to better video plans. Lastly, when online access to the environment is available, \method executes the current video plan and collects additional successful trajectories to further finetune the video generation model. A visual illustration of \method is shown in Figure~\ref{fig:framework}.

We first evaluate \method in two simulated robotic manipulation environments, Meta-World~\cite{yu2020meta} and iTHOR~\cite{kolve2017ai2}, and show that \method improves task success across all environments and tasks evaluated. Next, we provide a thorough study on the effect of different components in \method, including different types of feedback from the VLM, providing a recipe for utilizing VLM feedback for video generation. Lastly, we illustrate that \method can iteratively improve real-robot videos, providing early signal that robotics can be an important mean to ground video generation models in the real world.

\section{Background}
\label{background}

In this section, we provide the background on video generation as policy in a decision making process~\cite{du2023video}. We also introduce consistent diffusion models~\cite{song2023consistency,heek2024multistep, daras2024condiff}, which \method builds upon for self-refinement.

\subsection{Video as policy in sequential decision making}
We consider a predictive decision process similar to  \citet{du2024learning}: $\mathcal{P} := \langle\mathcal{X}, \mathcal{G}, \mathcal{A}, H, \mathcal{E}, \mathcal{R}\rangle$, where $\mathcal{X}$ denotes an image-based observation space, $\mathcal{G}$ denotes textual task description space, $\mathcal{A}$ denotes a low-level motor control action space, and $H\in\mathbb{R}$ denotes the horizon length. We denote $\pi(\cdot|x_0, g): \mathcal{X}\times\mathcal{G}\mapsto\Delta(\mathcal{X}^H)$\footnote{We use $\Delta(\cdot)$ to denote a probability simplex function} as the language conditioned video generation policy, which models the probability distribution over $H$-step image sequences $\mathbf{x}=[x_0, ..., x_H]$ determined by the first frame $x_0$ and the task description $g$. Intuitively, $\mathbf{x}\sim\pi(\cdot|x_0,g)$ correspond to possible visual paths for completing a task $g$. Given a sampled video plan $\mathbf{x}$, one can use a learned mapping $\rho(\cdot|\mathbf{x}):\mathcal{X}^H\mapsto\Delta\mathcal(\mathcal{A}^H)$ to extract motor controls from generated videos through a goal-conditioned policy~\cite{du2023video}, diffusion policy~\cite{black2023zero}, or dense correspondence~\cite{ko2023learning}. Once a sequence of motor controls $\mathbf{a}\in\mathcal{A}^H$ are extracted from the video, they are sequentially executed in the environment $\mathcal{E}$, after which a final reward $\mathcal{R}:\mathcal{A}^H\mapsto\{0, 1\}$ is emitted representing whether the task was successfully completed. For simplicity, we only consider finite horizon, episodic tasks. Given a previously collected dataset of videos labeled with task descriptions $\mathcal{D} = \{(\mathbf{x}, g)\}$, one can leverage behavioral cloning (BC)~\cite{pomerleau1988alvinn} to learn $\pi$ by minimizing
\begin{equation}
    \mathcal{L}_\text{BC}(\pi) = \E_{(\mathbf{x}, g)\sim\mathcal{D}}[-\log\pi(\mathbf{x}|x_0, g)].
    \label{eq:bc}
\end{equation}
Equation \ref{eq:bc} can be viewed as maximizing the likelihood of the videos in $\mathcal{D}$ conditioned on the initial frame and task description.

\subsection{Consistency Models}
Diffusion models~\cite{ho2020denoising,song2020score} have emerged as an important technique for generative modeling of high-dimensional data. 
During training, a diffusion model learns to map noisy data (at various noise levels) back to clean data in a single step. Concretely, let $x^{(0)}$ denote a clean image and $x^{(t)}$ denote the noisy image at noise level $t$, where $t\in [0, T]$, the training objective for a diffusion model $f_\theta(x^{(t)}, t)$ can be written as
\begin{equation}
    \mathcal{L}_\text{diffusion}(\theta) = \E_{x^{(0)},\epsilon,t}\left[\|f_\theta(x^{(t)}, t)-x^{(0)}\|^2\right],\label{eq:diffusion}
\end{equation}
where $\epsilon\in\mathcal{N}(0,I)$ is the added noise, and $x^{(t)} = \sqrt{\alpha_t}x^{(0)} + \sqrt{1 - \alpha_t}\epsilon$ where $\alpha_t$ are time-dependent noise levels. Although diffusion models have achieved high-quality image/video generation, they require hundreds or thousands of denoising steps during inference, which induces tremendous computational cost.
To overcome the slow sampling speed of diffusion models, \emph{consistency models}~\cite{song2023consistency,song2023improved} were initially proposed 
by enforcing a consistency loss across different noise levels, i.e.,
\begin{align}
    \mathcal{L}_\text{consistency}(\theta) &= \E_{x^{(0)},\epsilon,t_1,t_2}\Big[\|f_\theta(x^{(t_1)}, t_1) \nonumber \\
    &\quad-\texttt{stopgrad}\big(f_{\theta}(x^{(t_2)}, t_2)\big)\|^2\Big], \label{eq:consistency}
\end{align}
which encourages the output of the single-step map between different noise levels to be similar. In fact, both the diffusion loss in Equation~\ref{eq:diffusion} and the consistency loss in Equation~\ref{eq:consistency} can be understood as exploiting the structure of the denoising procedure which corresponds to an ordinary differential equation (ODE). Specifically, as introduced in~\cite{song2023consistency,song2020denoising}, the backward denoising procedure of a diffusion model can be characterized by an ODE, i.e.,
\begin{equation}\label{eq:ode}
    \frac{\mathrm{d}x^{(t)}}{\mathrm{d}t} = - t \cdot s(x^{(t)}, t),
\end{equation}
with $s(x^{(t)}, t)$ is some score function. During the entire path along $t\in (\epsilon, \infty]$, following this ODE should always map $x^{(t)}$ to $x^{(0)}$. If we parametrize the model $f(x^{(t)}, t)$ as the simulation following the ODE governed by $s(x^{(t)}, t)$, we obtain the diffusion loss~(\ref{eq:diffusion}). Meanwhile,  for all $t, t'\in (\epsilon, \infty]$, we have $f(x^{(t)}, t) = f(x^{(t')}, t')$ along the simulation path, which induces the consistency loss~(\ref{eq:consistency}). 
Therefore, we can combine the diffusion loss and the consistency loss together for model training, i.e., 
\begin{equation}
    \mathcal{L}(\theta) = \mathcal{L}_\text{diffusion}(\theta) + \lambda \cdot \mathcal{L}_\text{consistency}(\theta),
\end{equation}
where $\lambda$ denotes consistency regularization hyperparameter across different noise levels.

\section{Video Generation as an Agent}
\label{method}

In this section, we introduce \method, a framework for improving video plan generation. In Section~\ref{sec:consistency}, we develop \emph{self-conditioning consistency} to iteratively refine generated video plans. In Section~\ref{sec:vlm}, we describe how a diffusion model trained with self-conditioning consistency can leverage inference-time compute to refine generated video plans. Finally, in Section~\ref{sec:online}, we illustrate how \method closes the self-improvement loop by collecting additional online data to further enhance video generation and refinement.

\subsection{Refinement through self-conditioning consistency}\label{sec:consistency}

We consider first-frame-and-language conditioned video generation following~\cite{du2023video,ko2023learning}, which generates a sequence of image frames to complete the task described by the language starting from the initial image. In practice, generated videos often contain hallucinations ~\cite{yang2023learning}. While such inaccuracies may prevent a video plan from fully completing the task, the generated video may still make meaningful progress towards completing the task. Therefore, instead of independently sampling many videos and hoping that one may be free from hallucinations, we propose to refine previously generated videos iteratively.



\begin{figure}[t]
  \centering
  \includegraphics[width=0.98\linewidth]{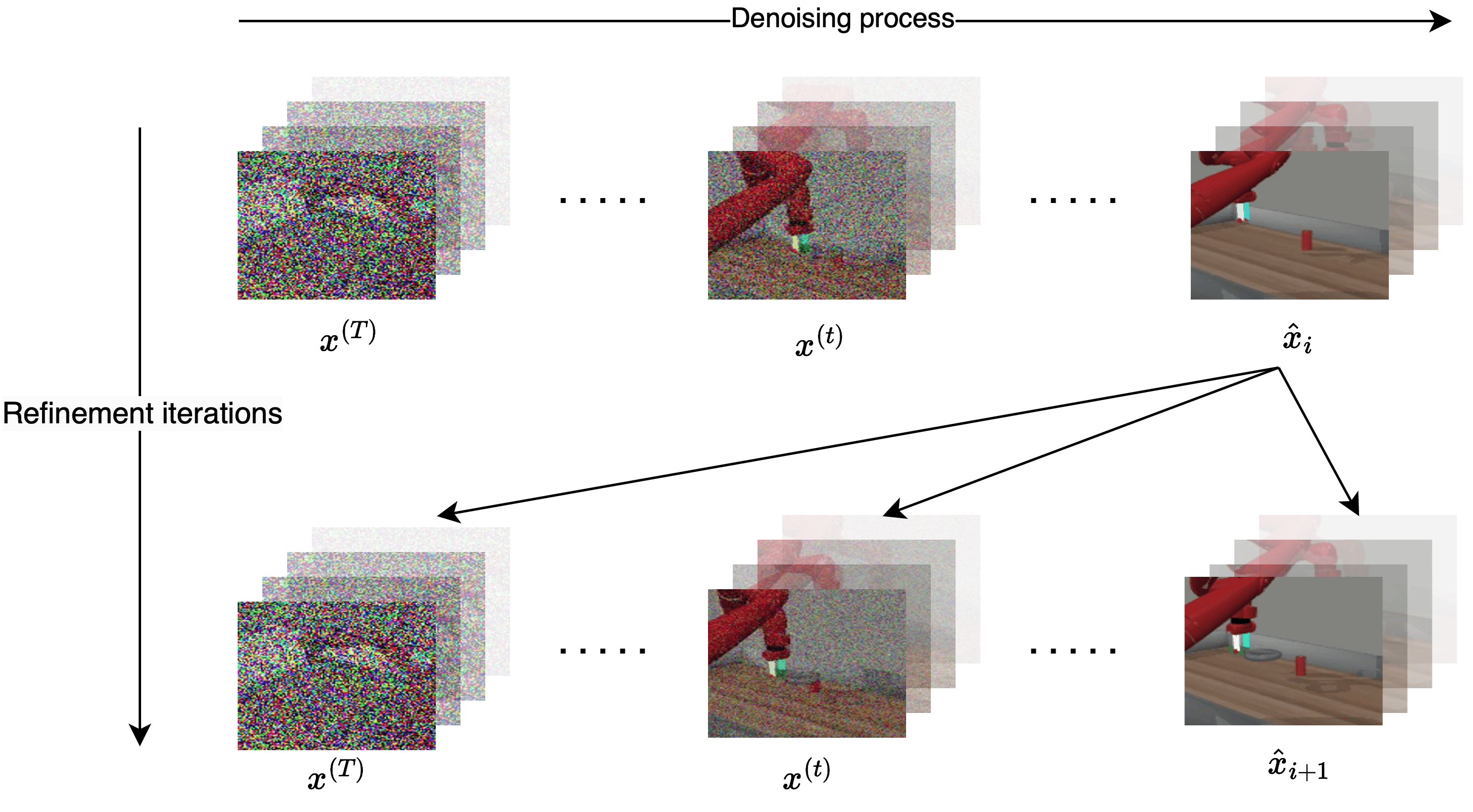}
  \caption{\textbf{An illustration of Self-Conditioning Consistency.} The horizontal direction represents the regular denoising process. The two rows represent two refinement iterations. $\hat{\mathbf{x}}_i$ denotes the generated video plan at refinement iteration $i$. We condition the refinement iteration $i+1$ on the generated video from the previous iteration $\hat{\mathbf{x}}_{i}$.}
  \label{fig:self_conditioned_consistency}
\end{figure}

Specifically, let $\mathbf{x}^{(0)}$ denote the ground truth video and $\hat{\mathbf{x}}$ a generated video from the original text-to-video diffusion model. We introduce a \emph{self-conditioning consistency} model, $\hat{f}_\theta(\hat{\mathbf{x}}, \mathbf{x}^{(t)}, t)$, which takes a generated video $\hat{\mathbf{x}}$ and a noisy version of the ground truth $\mathbf{x}^{(t)}$ as inputs to predict the clean video. This formulation enables iterative refinement by conditioning the model on its previous predictions, as illustrated in Figure~\ref{fig:self_conditioned_consistency}. We denote video samples from the refinement model after the $i$-th iteration as $\hat{\mathbf{x}}_{i}$.
Self-conditioning is inspired by a reparameterization of the implicit ODE solver for Equation \ref{eq:ode} ~\cite{song2020denoising,lu2022dpm,zhang2022fast, chen2022analog}. For instance,~\cite{song2020denoising} considered the first-order ODE solver for Equation~\ref{eq:ode} following:
\begin{equation}\label{eq:fast_ode_solver}
\mathbf{x}^{(t-1)} =  \sqrt{\alpha_{t-1}}\mathbf{x^{(0)}} + \sqrt{1 - \alpha_{t-1} - \sigma_t^2}\cdot s({\mathbf{x}^{(t)}, t}).     
\end{equation}

In \method, we adapt Equation \ref{eq:fast_ode_solver} by replacing the \(\mathbf{x}^{(0)}\) term with \(\mathbf{\hat{x}}\), the previously generated video sample, as illustrated in Figure~\ref{fig:self_conditioned_consistency}. In standard DDIM-based methods~\cite{song2020denoising}, \(\mathbf{x}^{(0)}\) is typically obtained as an intermediate estimate from \(\mathbf{x}^{(t)}\) within the \emph{same} iteration. In contrast, our approach reuses \(\mathbf{\hat{x}}\) from a \emph{previous} iteration, allowing for a self-conditioning mechanism that improves temporal coherence. 
By enforcing consistency across iterations, our method enables the denoising process to correct potential failures more effectively.

We learn the ODE solver through self-conditioning consistency by directly predicting the clean video $\mathbf{\hat{x}_{i+1}}$ using:
\begin{align}
    \mathcal{L}_\text{SC-consistency} (\theta) &= \E_{\hat{\mathbf{x}}, \mathbf{x}^{(0)}, t}\Big[\|\hat{f}_\theta(\hat{\mathbf{x}}, \mathbf{x}^{(t)}, t)-\mathbf{x}^{(0)}\|^2\Big] \nonumber\\ 
    &\quad + \mu \E_{\hat{\mathbf{x}}_1, \hat{\mathbf{x}}_2, t}\Big[\|\hat{f}_\theta(\hat{\mathbf{x}}_1, \mathbf{x}^{(t)}, t) \nonumber \\
    &\quad - \hat{f}_\theta(\hat{\mathbf{x}}_2, \mathbf{x}^{(t)}, t)\|^2\Big].\label{eq:video-refinement}
\end{align}

The first term in Equation~\ref{eq:video-refinement} represents the standard diffusion loss with the additional conditioning on $\hat{\mathbf{x}}$, while the second term regularizes the similarity between different refinement iterations (\(\mathbf{\hat{x}}_1\) and \(\mathbf{\hat{x}}_2\)) to promote coherence across iterations. This iterative refinement process distinguishes self-conditioning consistency from traditional consistency models. Combined with the standard objective for video diffusion:
\begin{equation}
    \mathcal{L}_\text{video-diffusion}(\theta) = \E_{\mathbf{x}^{(0)},\epsilon,t}\left[\|f_\theta(\mathbf{x}^{(t)}, t)-\mathbf{x}^{(0)})\|^2\right],\label{eq:video-diffusion}
\end{equation}
the overall objective for training a self-conditioning-consistent video diffusion model thus becomes:
\begin{equation}\label{eq:sc_loss}
    \mathcal{L}(\theta) = \mathcal{L}_\text{video-diffusion}(\theta) + \lambda \mathcal{L}_\text{SC-consistency}(\theta).
\end{equation}

Note that while the video generation model \(f_\theta\) and the video refinement model \(\hat{f}_\theta\) have different input arguments (first frame versus previously generated video), we can share their parameters to train a single unified model for both video generation and refinement tasks. This parameter-sharing approach allows us to leverage the same model architecture for generating initial video plans and iterative refinement. The training process for \(f_\theta\) and \(\hat{f}_\theta\) is detailed in Algorithm~\ref{alg:video_training} in Appendix~\ref{app:algo}.

\paragraph{Feedback Guided Self-Conditioning Consistency.} While we can refine videos only from previously generated samples, it may be desirable to condition the refinement process on any additional feedback for the previously generated video that is available (e.g., feedback from humans or vision language models critiquing which part of the generated video is unrealistic). 
When such feedback is available, we can have the refinement model $\hat{f}$ further take the additional feedback as input, combined with the task description, to guide the refinement process, i.e., 
\begin{equation}
\hat{f}_\theta(\mathbf{x}, \mathbf{x}^{(t)}, t| \text{feedback}), \label{eq:feedback}
\end{equation}
which can be plugged into our framework for learning using Equation~\ref{eq:sc_loss}.

\subsection{Inference through VLM guided video generation.}\label{sec:vlm}

After training the video generation model $f_\theta$ and the video refinement model $\hat{f}_\theta$ described in Equation~\ref{eq:video-diffusion} and Equation~\ref{eq:video-refinement}, 
we can sample from $f_\theta$ and iteratively apply $\hat{f}_\theta$ for video refinement. Specifically, let $\eta$ be the step size for the noise schedule, $\sigma_t$ be a time dependent noise term, 
\method first generates a video plan through
\begin{equation}
    \mathbf{x}^{(t-1)} = \mathbf{x}^{(t)} - \eta \cdot \nabla_\theta f_\theta(\mathbf{x}^{(t)}, t) + \sigma_t \cdot \epsilon.\label{eq:sampling}
\end{equation}
The sample $\hat{\mathbf{x}}$ after $T$ denoising steps corresponds to the generated video. Next, we can iteratively apply $\hat{f}_\theta$ to refine the generated video sample
\begin{equation}
    \hat{\mathbf{x}}_{i+1} = \hat{f}_\theta(\hat{\mathbf{x}}_{i}, \mathbf{x}^{(t)}, t),
\end{equation}
where $i$ denotes the video refinement iteration, with $\hat{\mathbf{x}}_0 = \hat{\mathbf{x}} = \mathbf{x}^{(T)}$. We denote the final video after refinement as $\hat{\mathbf{x}}_\text{refined}$. A natural question is when to stop the iterative video refinement process. We use a VLM as a proxy for the environment's reward to assess whether a refined video is likely to lead to successful execution in the environment. Specifically, we denote a VLM as $\hat{\mathcal{R}}$, which takes a refined video $\hat{\mathbf{x}}_{i}$ and returns a binary value $\{0, 1\}$ to determine whether a video is acceptable based on overall coherence, adherence to physical laws, and task completion (See prompt for VLM in Appendix~\ref{app:vlm_prompt}). With $\hat{\mathcal{R}}$, the refinement stops when the VLM decides that the refined video is acceptable. Namely, we have
\begin{equation}
    \hat{\mathbf{x}}_\text{refined} = \hat{\mathbf{x}}_{i^*}, \quad \text{where} \quad i^* = \min \left\{ i : \hat{\mathcal{R}}(\hat{\mathbf{x}}_{i}) = 1 \right\}
\end{equation}
Algorithm~\ref{alg:video_inference} in Appendix~\ref{app:algo} shows details for video plan generation, refinement, and selection during inference.
\begin{table*}[t!]
\centering
\setlength{\tabcolsep}{3pt}
\caption{\textbf{Meta-World Results.} The mean success rates of baselines and \method on 11 simulated robot manipulation environments from Meta-World. \method consistently outperforms baselines across all tasks.}
\scalebox{1}{
\begin{tabular}{l c c c c c c}
\toprule
& door-open & door-close & basketball & shelf-place & btn-press & btn-press-top \\
\midrule
AVDC & 30.7\% & 28.0\% & 21.3\% & 8.0\% & 34.7\% & 17.3\% \\ 
AVDC-Replan & 72.0\% & 89.3\% & 37.3\% & 18.7\% & 60.0\% & 24.0\% \\ 
\midrule
VideoAgent & 40.0\% & 29.3\% & 13.3\% & 9.3\% & 38.7\% & 18.7\% \\ 
VideoAgent-Online (Iter1) & 48.0\% & 40.0\% & 24.0\% & 12.0\% & 42.7\% & 36.0\% \\ 
VideoAgent-Online (Iter2) & 58.7\% & 50.7\% & 28.0\% & 18.7\% & 53.3\% & 41.3\% \\ 
VideoAgent-Online-Replan & \textbf{82.7\%} & \textbf{97.3\%}& \textbf{40.0\%} & \textbf{26.7\%} & \textbf{73.3\%} & \textbf{44.0\%}\\ 
\toprule
& faucet-close & faucet-open & handle-press & hammer & assembly & \cellcolor{gray!20}\textbf{Overall} \\
\midrule
AVDC & 12.0\% & 17.3\% & 41.3\% & 0.0\% & 5.3\% & \cellcolor{gray!20}19.6\% \\ 
AVDC-Replan & 53.3\% & 24.0\% & 81.3\% & 8.0\% & 6.7\% & \cellcolor{gray!20}43.1\% \\ 
\midrule
VideoAgent & 46.7\% & 12.0\% & 36.0\% & 0.0\% & 1.3\% & \cellcolor{gray!20}22.3\% \\ 
VideoAgent-Online (Iter1) & 53.3\% & 28.0\% & 52.0\% & 1.3\% & 5.3\% & \cellcolor{gray!20}31.2\% \\ 
VideoAgent-Online (Iter2) & 58.7\% & 36.0\% & 64.0\% & 1.3\% & 9.3\% & \cellcolor{gray!20}38.2\% \\ 
VideoAgent-Online-Replan & \textbf{74.7\%} & \textbf{46.7\%} & \textbf{86.7\%} & \textbf{8.0\%} & \textbf{10.7\%} & \cellcolor{gray!20}\textbf{53.7\%}\\ 
\bottomrule
\end{tabular}
}
\label{tab:mw_results}
\end{table*}

\subsection{Self-improvement through online finetuning}\label{sec:online}
In addition to video refinement through self-conditioning consistency and inference-time compute as described in Section~\ref{sec:consistency} and Section~\ref{sec:vlm}, we can further characterize the combination of video generation and video refinement as a policy, which can be improved by training on additional data collected from the environment during online interaction. Specifically, the goal is to maximize the expected returns of a policy through trial-and-error interaction with the environment:
\begin{equation}
    \mathcal{J}_\text{online}(\theta) = \E\left[\mathcal{R}(\mathbf{a})\,|\,\pi_\theta, \rho, \mathcal{E}\right],\label{eq:online}
\end{equation}
where $\mathcal{R}$ is the true reward function, $\mathcal{E}$ is the interactive environment, and $\pi_\theta$ corresponds to the combination of $f_\theta$ and $\hat{f}_\theta$.

A broad array of reinforcement learning methods~\cite{sutton2018reinforcement} such as policy gradient~\cite{Schulman2017ProximalPO} can be employed to maximize the objective in Equation~\ref{eq:online}. For simplicity, we consider the setup of first executing the policy in the environment, then filtering for successful trajectories, continuing finetuning the video policy using additional online data, and executing the finetuned policy again to collect more data. Specifically, each online iteration constructs an additional dataset by rolling out the policy $\pi_\theta$ at the current online iteration
\begin{equation}
    \mathcal{D}_\text{new} = \left\{ \hat{\mathbf{x}}_\text{refined} \sim \pi_\theta(x_0, g) \mid \mathcal{R}(\rho(\hat{\mathbf{x}}_\text{refined})) = 1 \right\},
\end{equation}
where $\rho$ is the optical flow model that maps the refined video to low-level control actions. See Algorithm~\ref{alg:online_finetuning} in Appendix~\ref{app:algo} for details of online policy finetuning.

\section{Experiments}
\label{experiments}

We now evaluate the performance of \method. We introduce the experimental settings and variants of \method in Section \ref{sec:exp_setup}, measure end-to-end success rate of \method against the baselines in Section~\ref{sec:metaworld}, and study the effect of different components of \method in Section~\ref{sec:exp_ablation}. Finally, we show that \method is effective in improving the quality of real robotic videos in Section~\ref{sec:realworld}.

\subsection{Datasets and Experimental Setups}\label{sec:exp_setup}

\paragraph{Datasets and Environments.} We follow the same evaluation setting as \citet{ko2023learning}, which considers three datasets: Meta-World~\cite{yu2020meta}, iTHOR~\cite{kolve2017ai2}, and BridgeData V2~\cite{walke2023bridgedata}. Meta-World consists of 11 robotic manipulation tasks performed by a simulated Sawyer arm, with video demonstrations captured from three distinct camera angles. iTHOR is a simulated 2D object navigation benchmark, where an agent searches for specified objects across four room types. BridgeData V2 is a real-world dataset of robotic manipulation. See more details of datasets and environments in Appendix~\ref{app:Data}.

\paragraph{Baselines and \method Variants.} We consider the following methods for comparison:
\begin{itemize}[leftmargin=*,topsep=0pt]
\setlength{\itemsep}{0mm}
\item \textbf{AVDC} (baseline). This is the Actions from Video Dense Correspondences~\cite{ko2023learning} baseline, which synthesizes a video and predicts optical flow to infer actions.
\item \textbf{AVDC-Replan} (baseline). When the movement stalls,
AVDC-replan re-runs video generation and action extraction from the flow model to execute a new plan.
\item \textbf{\method}. Our proposed video refinement model through self-conditioning consistency as introduced in Section~\ref{sec:consistency}. 
\method generates video and iteratively refines a video plan. We use GPT-4 Turbo for selecting the best video plan during inference (Section~\ref{sec:vlm}).
\item \textbf{\method-Online}. As actions are executed in the online environment,
successful trajectories are collected and used to continue training the video generation and refinement model, as described in Section \ref{sec:online}.
\item \textbf{\method-Replan}. This variant incorporates online filtering of successful trajectories with the replanning mechanism, where replanning is conducted first, and more successful trajectories after replanning are added back to the training data.
\end{itemize}

\subsection{End-to-End Task Success}\label{sec:metaworld}

\paragraph{Meta-World.} We report the task success rates of baselines and \method in Table~\ref{tab:mw_results}. Following~\cite{ko2023learning}, we evaluate performance across three camera poses with 25 seeds per pose. Without online environment access, \method improves the overall success rate through self-conditioning consistency alone from 19.6\% (AVDC) to 22.3\%. For certain difficult tasks, e.g., faucet-close, \method improves performance from 12.0\% to 46.7\%. With online data collection, \method-Online further improves success rates with each additional online iteration of rolling out the policy, collecting successful trajectories, and finetuning. \method-Online can be further combined with replanning, achieving 53.7\% overall success, surpassing prior state-of-the-art on this benchmark. Detailed baseline comparisons are provided in Appendix~\ref{abl:baselines}, and qualitative improvements in refined videos are shown in Figure~\ref{fig:metaworld_qual} in Appendix~\ref{abl:examples}.

\paragraph{iTHOR.} 
\begin{table}[t]
\vspace{-3mm}
\centering
\small\setlength{\tabcolsep}{3pt}
\captionof{table}{\textbf{iThor Success Rates} comparing VideoAgent with the AVDC baseline.}
\label{tab:thor_results}
\begin{tabular}{l c c}
\toprule
    Room & AVDC & VideoAgent \\
    \midrule
    Kitchen & 26.7\% & \textbf{28.3\%} \\ 
    Living Room & 23.3\% & \textbf{26.7\%} \\ 
    Bedroom & 38.3\% & \textbf{41.7\%} \\ 
    Bathroom & 36.7\% & \textbf{40.0\%} \\ 
    \midrule
    \rowcolor{gray!20} Overall & 31.3\% & \textbf{34.2\%} \\
    \bottomrule
\end{tabular}
\vspace{-4mm}
\end{table}

Next, we evaluate \method on iThor. Due to the high computational cost of running the iThor simulator, we focus only on evaluating self-conditioning consistency (without online access). We follow the same setup as \citet{ko2023learning}, where we measure the average success rate across four rooms each with three objects using 20 seeds. As shown in Table \ref{tab:thor_results}, VideoAgent consistently outperforms the baseline, demonstrating the effectiveness of self-conditioning consistency in producing more plausible video plans.

\subsection{Effect of Different Components in \method} \label{sec:exp_ablation}

In this section, we aim to understand the effect of different components of \method. Specifically, we focus on the effect of (1) different types of feedback given to the refinement model, (2) the number of refinement and online iterations, and (3) the quality of the VLM feedback.

\subsubsection{Effect of Different VLM Feedback}
\begin{table}[t]
\centering
\small\setlength{\tabcolsep}{3pt}
\captionof{table}{\textbf{Effect of Different Feedback} used to train the refinement model. Descriptive feedback from the VLM leads to higher improvement in task success.}
\label{tab:mw_feedback_results}
\begin{tabular}{l c}
\toprule
    & \textbf{Overall} \\
    \midrule
    AVDC & 19.6\% \\ 
    \midrule
    VideoAgent & 22.3\% \\ 
    VideoAgent-Binary & 23.8\% \\ 
    VideoAgent-Suggestive & 26.6\% \\ 
    \bottomrule
\end{tabular}
\end{table}

In the previous section, we only used VLM during inference to determine when to stop refining a generated video. However, it is natural to wonder if information-rich feedback from the VLM, such as language descriptions of which part of a generated video to improve, might lead to better refined videos. To answer this question, we propose a few variants of \method according to the feedback available when training the video refinement model as in Equation~\ref{eq:feedback}. Specifically, we use \method to denote training the video refinement model only conditioned on the original task description. \method-Binary denotes additionally conditioning on whether a generated video is determined to be successful by the VLM. \method-Suggestive denotes conditioning additionally on language feedback from the VLM on which part of the video needs improvement and how the video can be improved. We train these three versions of the video refinement model, and report the overall task success from Meta-World in Table \ref{tab:mw_feedback_results}. We see that VideoAgent-Binary improves upon the base VideoAgent, while training with descriptive feedback in VideoAgent-Suggestive leads to even better performance. This suggests that richer feedback from the VLM can facilitate better training of the video refinement model. Improvement for each individual task can be found in the Appendix \ref{abl:vlm_feedback}.

\subsubsection{Effect of the Number of Iterations.}
\begin{figure*}[t]
    \centering
    \begin{minipage}[t]{0.48\textwidth}
        \centering
        \includegraphics[width=\textwidth]{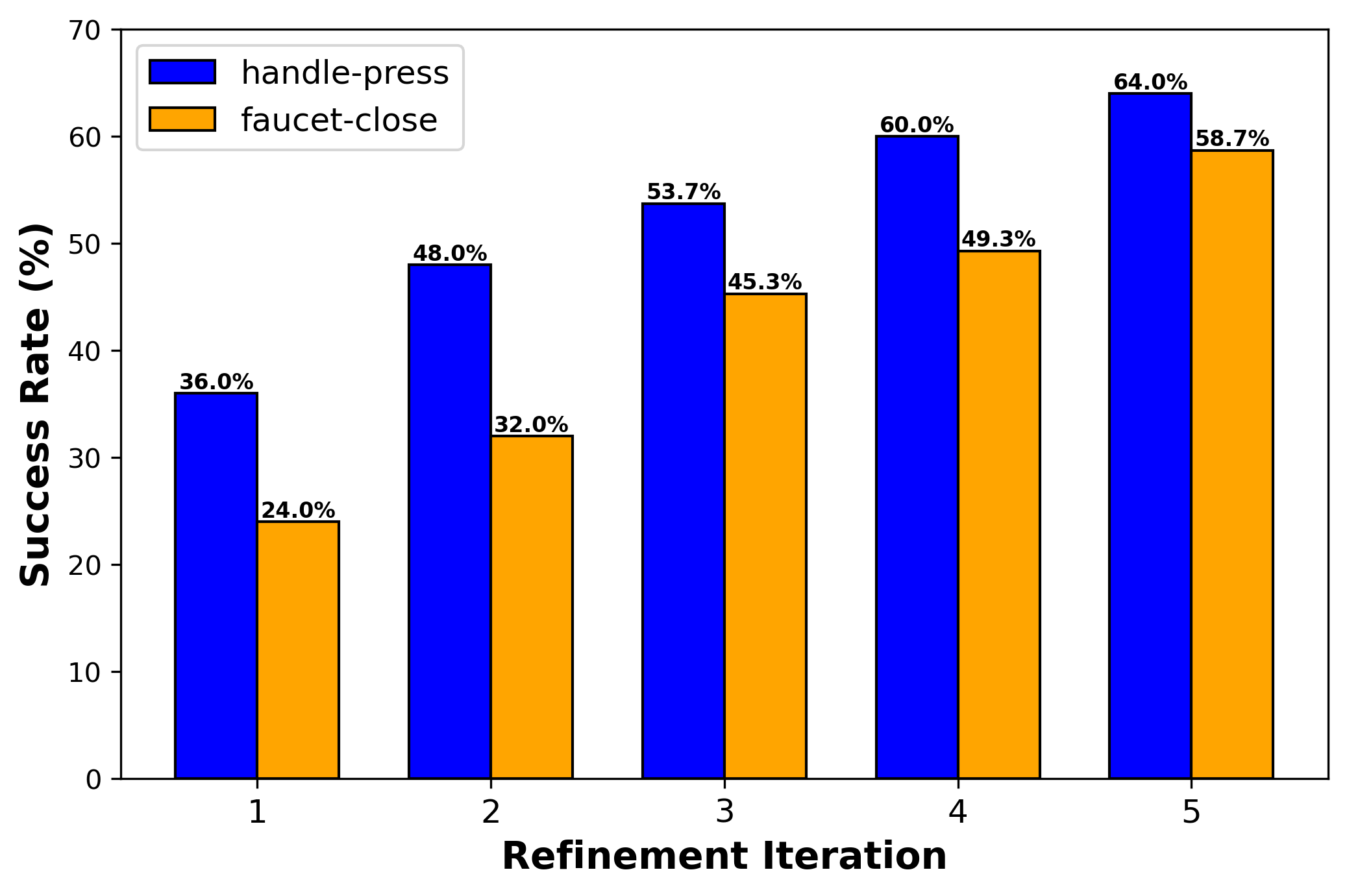}
        \caption{\textbf{Effect of Refinement Iterations.} The accuracy of downstream tasks generally increases as the number of refinement iteration increases.}
        \label{fig:iterative_refinement}
    \end{minipage}%
    \hfill
    \begin{minipage}[t]{0.48\textwidth}
        \centering
        \includegraphics[width=\textwidth]{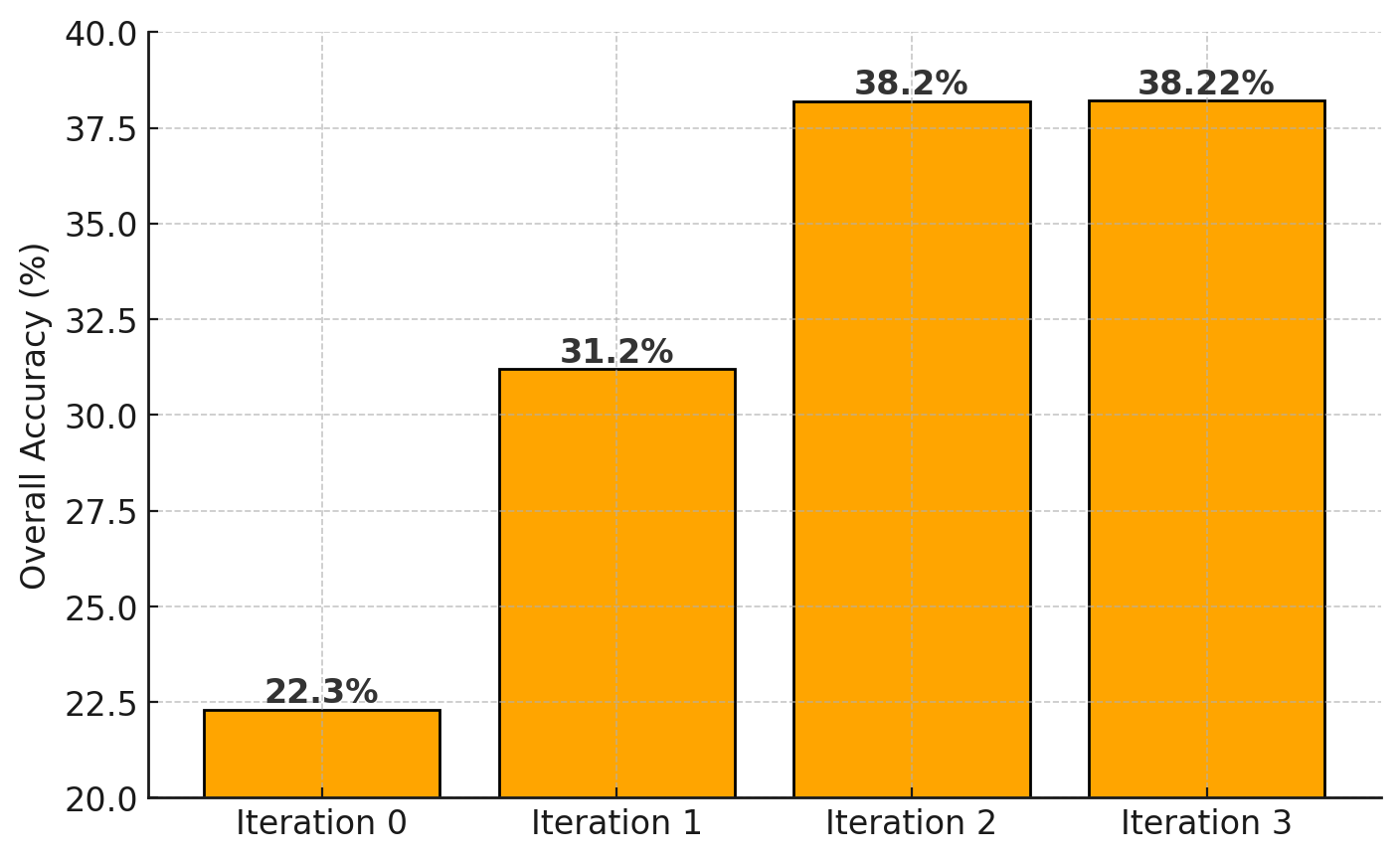}
        \caption{\textbf{Effect of Online Iterations.} The overall task success of \method increases as the number of online iterations increases.}
        \label{fig:accuracy_plot}
    \end{minipage}%
\end{figure*}

Next, we want to understand whether more refinement iterations and online finetuning iterations lead to higher task success. We found that while different tasks require a different number of iterations to achieve the best performance, \method does perform better as the number of refinement and online iterations increases, as shown in Figure~\ref{fig:iterative_refinement} and Figure~\ref{fig:accuracy_plot}. During video refinement, specific tasks such as handle-press and faucet-close continue to show improvement even at the fifth refinement iteration. Faucet-close especially benefits from more refinement iterations, bringing success rate from 24.0\% to 58.7\% after five refinement iterations. The improved task success rates across refinement and online iterations suggests that self-conditioning consistency discussed in Section~\ref{sec:consistency} and online interaction discussed in Section~\ref{sec:online} can indeed effectively reduce hallucination and improve physical plausibility in the generated videos.




\subsubsection{Accuracy of VLM feedback}
\begin{table}[t]
\vspace{-3mm}
\centering
\small\setlength{\tabcolsep}{2pt}
\captionof{table}{\textbf{VLM Performance} measured according to whether a VLM considers a generated video as acceptable using human label as the ground truth.}
\label{tab:vlm_acc}
\begin{tabular}{ccccc}
\toprule
& \textbf{Precision} & \textbf{Recall} & \textbf{F1-Score} & \textbf{Accuracy} \\
\midrule
\textbf{Unweighted } & 0.65 & 0.89 & 0.76 & 0.69 \\
\textbf{Weighted } & \textbf{0.92} & 0.58 & 0.71 & 0.75 \\
\textbf{Without Cam 3} & 0.91 & 0.71 & 0.80 & 0.79 \\
\bottomrule
\end{tabular}
\vspace{-3mm}
\end{table}
Since this work is among the first to leverage a VLM to give feedback for video generation, it is crucial to understand whether a VLM can in fact achieve a reasonable accuracy in providing feedback for video generation. To quantify the performance of a VLM, we use human labels on whether a generated video is acceptable as the ground truth, and measure precision, recall, F1-score, and accuracy based on whether GPT-4 Turbo thinks the generated video is acceptable according to trajectory smoothness (consistent across sequential frames), physical stability, and achieving the goal (See full prompt in Appendix~\ref{app:vlm_prompt}). We report the average result across 
36 generated videos from 
the Meta-World dataset in Table~\ref{tab:vlm_acc}. We see that the original prompt we used (Unweighted) achieves 69\% accuracy, suggesting that the VLM is able to somewhat judge generated videos but not always accurately. Since \method uses multiple refinement iterations, we want to avoid false positives where a bad video is accidentally accepted. We can achieve this by penalizing false positives through reweighing its cost in the prompt, which leads to the VLM rejecting videos when the VLM is uncertain about the video's acceptability. This adjustment results in a significant increase in precision as shown in Table~\ref{tab:vlm_acc}. This weighted version of the prompt is used in the experiments in Section~\ref{sec:metaworld}.

\paragraph{Partial Observability.} In the AVDC experimental setup, center cropping the third camera (what is used in the pipeline) often results in most of the robot arm being outside of the frame. We found that the accuracy of the VLM is affected by such partial observability. As shown in Table~\ref{tab:vlm_acc}, removing the third camera from the prompt leads to much higher accuracy. 

\paragraph{Descriptive Feedback.} While VLM can provide binary feedback on whether a generated video is acceptable, we also measure the accuracy of the VLM in giving more descriptive feedback such as identifying the issue and providing suggestions on how to improve the video. We use three examples with human written language feedback as prompt for in-context learning. GPT-4 Turbo achieves 73.5\% accuracy on identification and 86.1\% accuracy on suggestion, as evaluated by humans. This result is highly encouraging and opens up future directions of leveraging descriptive feedback from VLMs to improve video generation.



\subsection{Evaluating Self-Refinement on Real-World Videos} \label{sec:realworld}


In this section, we evaluate \method's ability to refining real-world videos, which often contain higher variability, intricate details, nuanced behaviors, and complex interactions. We study the effect of video refinement using both quantitative metrics and qualitatively for holistic evaluation.

\begin{table}[t]
    \centering
    \caption{\textbf{BridgeData-V2 Results.} Quantitative metrics comparing AVDC and \method on generated Bridge data. \method outperforms the baseline according to all except for one metric.}
    \small
    \begin{tabular}{l c c}
        \toprule
        \textbf{Metrics} &\textbf{AVDC} & \textbf{Video Agent} \\
        \midrule
        \rowcolor{gray!20} \textbf{Clip Score} &22.39 & \textbf{22.90} \\
        \rowcolor{gray!20} \textbf{Flow Consistency} & 2.48 ± 0.00 & \textbf{2.59 ± 0.01} \\
        \midrule 
        Visual Quality & 1.97 ± 0.003 & \textbf{2.01 ± 0.003}\\
        Temporal Consistency & 1.48 ± 0.01 & \textbf{1.55 ± 0.01} \\
        Dynamic Degree & \textbf{3.08 ± 0.01} &  3.07 ± 0.02\\
        Text to Video Alignment & 2.26 ± 0.003 &  \textbf{2.30 ± 0.03} \\
        Factual Consistency & 2.02 ± 0.004 &  \textbf{2.07 ± 0.01} \\

        \midrule
        \rowcolor{gray!20}\textbf{Average Video Score} & 2.16 ± 0.01 & \textbf{2.20 ± 0.01} \\
        \midrule
        \rowcolor{gray!20} \textbf{Human Eval on Task Success} & 42.0\%& \textbf{64.0\%} \\
        \bottomrule
    \end{tabular}
    \label{table:bridge_metrics_comparison}
\end{table}

\begin{figure*}[t]
    \centering
    \includegraphics[width=0.94\linewidth]{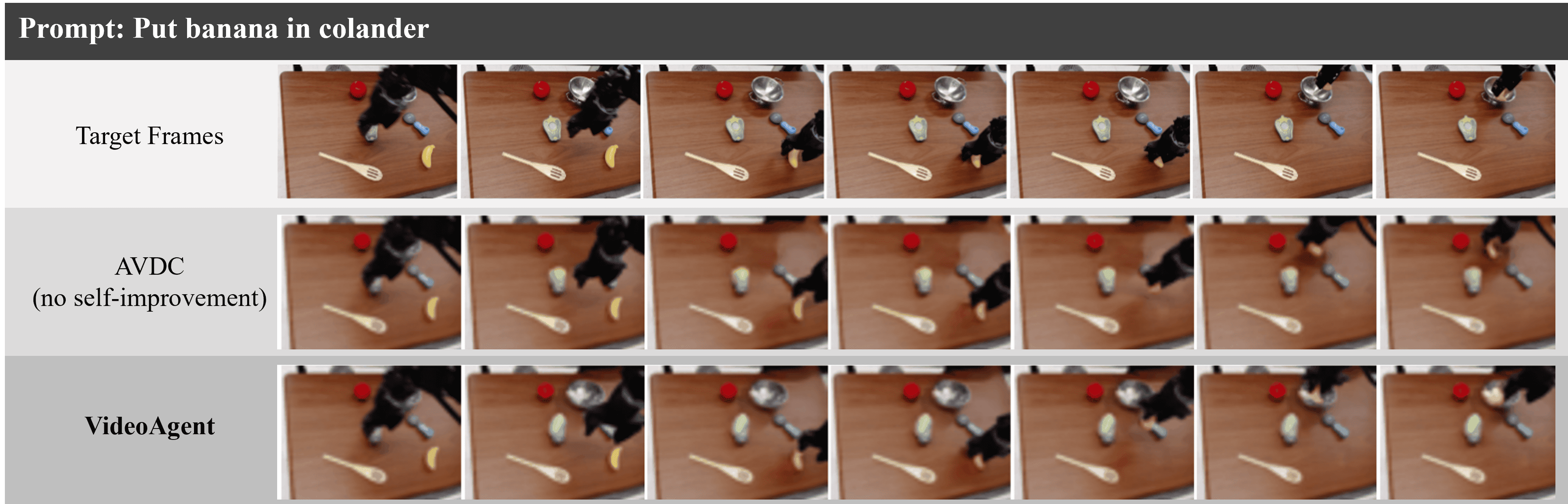}
    \caption{\textbf{Correcting Hallucinations in Video Generation:} The AVDC model hallucinates after the second frame, removing the colander and placing the banana on the table. In contrast, VideoAgent accurately retains the colander's position and correctly places the banana inside.}
    \label{fig:bridge_qual}
\end{figure*}
\paragraph{Quantitative Evaluation.} Following previous literature on video generation, we consider two reference-free metrics, CLIP Score~\cite{hessel2021clipscore} and Flow Consistency~\cite{teed2020raft}, as well as a set of Video-Scores~\cite{he2024videoscore}. CLIP Score measures the cosine similarity between frame feature and text prompt, whereas Flow Consistency measure the smoothness and coherence of motion in the videos calculated from the RAFT model. Video-Scores use five sub-metrics with a focus on correlation with human evaluation and real-world videos.

We report the average across 2250 
videos generated from the AVDC baseline and from \method in Table \ref{table:bridge_metrics_comparison}. \method performs better according to all metrics except for Dynamic Degree from Video-Score (which shows similar performance between the two methods). 
Notably, the gain is significant in metrics critical for instruction following and real-world videos, such as CLIP Score, Factual Consistency, and Text-to-Video Alignment. Improvement in Flow Consistency and Temporal Consistency suggests that VideoAgent produces smoother and more physically plausible videos that adhere better to the physical constraints of the real-world. This directly translates to better performance in real-world robotic tasks in Table~\ref{tab:mw_results}.

\paragraph{Qualitative Evaluation.} Next, we qualitatively evaluate generated videos from the AVDC baseline and from \method. We collect 50 generated videos from each model and conduct human evaluation on whether a generated video looks realistic. Videos with refinement from \method improves the acceptance rate by 22\% as shown in Table~\ref{table:bridge_metrics_comparison}. We further show an example video with and without refinement in Figure~\ref{fig:bridge_qual}, where the baseline (middle row) hallucinates (the bowl disappears) whereas \method produces the video that completes the task (bottom row). We also present a more fine-grained analysis of Visual Quality, Temporal Consistency, Dynamic Degree, Text to Video Alignment, and Factual Consistency evaluated by humans in the Appendix \ref{app: human_eval_finegrained} with the metrics in Table \ref{table:bridge_human_eval}, which further echos the results of human evaluations presented in Table \ref{table:bridge_metrics_comparison}.

\section{Conclusion and Future Work}

We have presented \method, where a video generation model acts as an agent by generating and refining video plans, converting video plans into actions, executing the actions in an environment, and collecting additional data for further self improvement. Through interaction with an external environment, \method provides a promising direction for grounding video generation in the real world, thereby reducing hallucination and unrealistic physics in the generated videos according to real-world feedback. In order to fully achieve this overarching goal, \method needs to overcome a few limitations, which calls for future work:
\begin{itemize}[leftmargin=*,topsep=0pt,itemsep=0pt]
    \item In the online setting, \method only considers filtering for successful trajectories for further finetuning. Exploring other algorithms such as online RL is interesting future work.
    \item \method utilizes optical flow for action extraction. It would be interesting to see how \method works with inverse dynamics model or image-goal conditioned diffusion policy.
    \item We only measured end-to-end task success in simulated robotic evaluation settings. It would be interesting to see how \method works with real robotic systems.
    \item As additional data is being collected in the online setting, in addition to finetuning the video prediction model, one can also finetune the action extraction module (flow model), and the VLM feedback model using the additionally collected data, which we defer to future work.
    \item \method trades off inference-time compute for better performance by iteratively refines generated video plans under the guidance of a VLM. Exploring other inference-time search strategies to further improve video sample quality is an interesting area of research.
\end{itemize}

\section{Impact Statement}

\method introduces a novel self-conditioning consistency mechanism that enables iterative refinement of generated video plans, significantly improving long-horizon task completion. By leveraging previously generated video segments for refinement, \method mitigates hallucinations and enhances temporal consistency without requiring extensive interaction with the environment. This reduces the need for costly and time-consuming data collection while still achieving state-of-the-art success rates in simulated robotic environments. Furthermore, \method’s ability to refine plans without relying on replanning makes it highly adaptable to real-world applications, including robotics, autonomous systems, and video-based reinforcement learning. This work advances scalable and generalizable video generation techniques, contributing to the broader goal of AI agents that can reason and act through visual understanding.

\nocite{langley00}

\bibliography{icml2025_conference}
\bibliographystyle{icml2025}

\clearpage
\newpage
\appendix
\onecolumn
\begin{center}
{\huge Appendix}
\end{center}



\section{Algorithms}\label{app:algo}
\begin{algorithm}[H]
\caption{Training of Video Generation and Refinement Models with VLM Feedback}
\label{alg:video_training}
\KwIn{Dataset $\mathcal{D}$, learning rate $\gamma$, total training iterations $N$, initial model parameters $\theta$, video generation model $f_\theta$, video refinement model $\hat{f}_\theta$, VLM $\hat{\mathcal{R}}$}

\For{iteration $= 1$ \KwTo $N$}{
    Sample $\{ (\mathbf{x}^{(0)}, g) \} \sim \mathcal{D}$ and $t \sim \text{Uniform}(\{0,1,\dots,T\})$\;
    Compute vanilla diffusion loss: \\
    \Indp
    $\mathcal{L}_\text{video-diffusion} = \left\| f_\theta(\mathbf{x}^{(t)}, t) - \mathbf{x}^{(0)} \right\|^2$\;
    \Indm
    Generate $\hat{\mathbf{x}}$ following Equation~\ref{eq:sampling} and sample $\texttt{feedback} \sim \hat{\mathcal{R}}(\cdot|\hat{\mathbf{x}})$\;
    Compute consistency loss: \\
    \Indp
    $\mathcal{L}_\text{SC-consistency} = \left\| \hat{f}_\theta(\hat{\mathbf{x}}, \mathbf{x}^{(t)}, t\,|\, \texttt{feedback}) - \mathbf{x}^{(0)} \right\|^2$\;
    \Indm
    Update parameters: \\
    \Indp
    $\theta \leftarrow \theta - \gamma \nabla_\theta \left( \mathcal{L}_\text{video-diffusion} + \mathcal{L}_\text{SC-consistency} \right)$\;
    \Indm
}
\end{algorithm}
\begin{algorithm}[H]
\caption{VLM Guided Replan}
\label{alg:video_inference}
\KwIn{Initial frame $x_0$, task description $g$, Reward $\mathcal{R}$, Environment $\mathcal{E}$, VLM $\hat{\mathcal{R}}$, $\text{max\_refine\_iterations}$, $\text{max\_replans}$}

\For{$\text{replan\_count} = 1$ \KwTo $\text{max\_replans}$}{
    $\hat{\mathbf{x}} \leftarrow \pi_{\theta}(x_0, g)$\;
    \For{$i = 0$ \KwTo $\text{max\_refine\_iterations}$}{
        $\text{response} \leftarrow \hat{\mathcal{R}}(\hat{\mathbf{x}}_{(i)}, g)$\;
        \textbf{if} $\text{response} == \text{ACCEPT}$ then \textbf{break}\;
        $\hat{\mathbf{x}}_{(i+1)} \leftarrow \pi_{\theta}(\hat{\mathbf{x}}_{(i)}, x_0, g)$\;
    }
    $\text{success} \leftarrow \mathcal{R}(\rho(\hat{\mathbf{x}}_\text{refined}))$\;
    \textbf{if} $\text{success}$ then \textbf{break}\;
    
    $x_0 \leftarrow \mathcal{E}.\text{get\_state()}$\;
}
\end{algorithm}

\begin{algorithm}[H]
\caption{Online Finetuning of Video Generation and Refinement Models}
\label{alg:online_finetuning}
\KwIn{Dataset $\mathcal{D}$, policy $\pi_{\theta}$, Reward $\mathcal{R}$, Environment $\mathcal{E}$}
\For{iteration $i = 1$ to $N$}{
    $\mathcal{D}_\text{new} \leftarrow \emptyset$\;
    \For{each $(\cdot, g)$ in $\mathcal{D}$}{
        $x_0 \leftarrow \mathcal{E}.\text{reset}(g)$\;
        $\hat{x}_\text{refined} \sim \pi_{\theta}(x_0, g)$\;
        \If{$\mathcal{R}(\rho(\hat{x}_\text{refined}))$}{
            $\mathcal{D}_\text{new} \leftarrow \mathcal{D}_\text{new} \cup (\hat{x}_\text{refined}, g)$\;
        }
    }
    $\mathcal{D} \leftarrow \mathcal{D} \cup \mathcal{D}_\text{new}$\;
    Finetune $\theta$ using Algorithm 1 on $\mathcal{D}$\;
}
\end{algorithm}

\section{Prompt Structure for VLM Feedback}\label{app:vlm_prompt}
\subsection{Binary Classification}
We employ a structured prompting strategy to provide feedback on video sequences for the zero-shot classification. The process consists of one Query-Evaluation Phase, each with distinct sub-goals.

\begin{promptbox}{BINARY CLASSIFICATION}
\textbf{Task:} You are a video reviewer evaluating a sequence of actions presented as seven consecutive image uploads, which together represent a single video. You are going to accept the video if it completes the task and the video is consistent without glitches.

\textbf{Query-Evaluation Phase:}
\begin{itemize}[leftmargin=*]
  \item \textbf{Inputs Provided:}
  \begin{itemize}[leftmargin=*]
    \item \textbf{Textual Prompt:} Describes the task the video should accomplish.
    \item \textbf{Conditioning Image:} Sets the fixed aspects of the scene.
    \item \textbf{Sequence of Images (7 Frames):} Represents consecutive moments in the video to be evaluated.
  \end{itemize}
  \item \textbf{Evaluation Process:}
  \begin{itemize}[leftmargin=*]
    \item \textbf{View and Analyze Each Frame:} Examine each image in sequence to understand the progression and continuity of actions.
    \item \textbf{Assess Overall Coherence:} Determine if actions transition smoothly and logically from one image to the next.
    \item \textbf{Check for Physical Accuracy:} Ensure adherence to the laws of physics, identifying any discrepancies.
    \item \textbf{Verify Task Completion:} Confirm the sequence accomplishes the task described in the textual prompt.
    \item \textbf{Identify Inconsistencies:} Detect inconsistencies in object movement or overlaps that do not match the conditioning image.
  \end{itemize}
  \item \textbf{Evaluation Criteria:}
  \begin{itemize}[leftmargin=*]
    \item Accept the sequence if it is a coherent video that completes the task.
    \item Reject the sequence if any frame fails to meet the criteria, showing inconsistencies or not achieving the task. Be very strict, rejecting even minor errors.
  \end{itemize}
  \item \textbf{Response Requirement:}
  \begin{itemize}[leftmargin=*]
    \item Provide a single-word answer: \textit{Accept} or \textit{Reject}. Do not give reasoning.
  \end{itemize}
  \item \textbf{Additional Notes:}
  \begin{itemize}[leftmargin=*]
    \item No further clarification can be requested.
    \item Elements from the conditioning image must match those in each frame of the sequence.
  \end{itemize}
\end{itemize}
\end{promptbox}

\subsection{Identification and Suggestion:}
We employ a structured prompting strategy to provide descriptive feedback on video sequences via an in-context few-shot classification setup. The process consists of one Query-Evaluation Phase, each with distinct sub-goals.
\begin{figure*}[ht]  
\centering
\begin{promptbox}{IDENTIFICATION AND SUGGESTION}
\textbf{Task:} You are a video reviewer tasked with evaluating a series of actions depicted through eight consecutive image uploads. These images together simulate a video. This task is structured as a few-shot learning exercise, where you will first review three examples and then apply learned principles to new queries.
\textbf{Query-Evaluation Phase:}
\begin{itemize}
  \item \textbf{Inputs Provided:}
  \begin{itemize}
    \item \textbf{Textual Prompt:} Describes the intended outcome or task the video aims to accomplish.
    \item \textbf{Conditioning Image:} Establishes the fixed elements of the scene.
    \item \textbf{Sequence of Images (7 Frames):} Illustrates consecutive moments in the video, representing the action sequence.

  \end{itemize}
  \item \textbf{Evaluation Process:}
  \begin{itemize}
    \item \textbf{Frame-by-Frame Analysis:} Carefully examine each of the seven images to understand the progression and continuity of actions.
    \item \textbf{Assess Overall Coherence:} Evaluate the sequence as a whole to determine if the actions transition smoothly from one frame to the next while maintaining logical progression.
    \item \textbf{Check for Physical Accuracy:} Ensure each frame complies with the laws of physics, identifying any discrepancies in movement or positioning.
    \item \textbf{Verify Task Completion:} Confirm if the sequence as a whole accomplishes the task described in the textual prompt.
    \item \textbf{Identify Inconsistencies:} Detect inconsistencies in object movement or overlaps that contradict the fixed scene elements depicted in the conditioning image.

  \end{itemize}
  \item \textbf{Evaluation Criteria:}
  \begin{itemize}
    \item \textbf{Descriptive Feedback:} Based on your evaluation, provide a concise, constructive sentence suggesting specific improvements. Focus on enhancing physical accuracy and task fulfillment based on identified inconsistencies or discrepancies.
    
  \end{itemize}
  \item \textbf{Response Requirement:}
  \begin{itemize}
    \item Feedback must be derived from your observations during the evaluation and not exceed 20 words.
  \end{itemize}
  \item \textbf{Additional Notes:}
  \begin{itemize}
    \item No further clarification can be requested.
    \item Elements from the conditioning image must match those in each frame of the sequence.
  \end{itemize}
\end{itemize}
\end{promptbox}
\end{figure*}

\section{Task Descriptions and In-Context Examples for VLM Feedback}\label{app:icl_example}

\begin{promptbox}{TASK DESCRIPTION AND SUCCESS CRITERIA}
\begin{itemize}
  \item \textbf{door-open}: The robot arm has to open the door by using the door handle.
  \item \textbf{door-close}: The robot arm has to close the door by pushing the door or the handle.
  \item \textbf{basketball}: The robot arm has to pick up the basketball and take it above the hoop.
  \item \textbf{shelf-place}: The robot arm has to pick up the blue cube and place it on the shelf.
  \item \textbf{button-press}: The robot arm has to press the red button from the side by pushing it inside.
  \item \textbf{button-press-topdown}: The robot arm has to press the red button from the top by pushing it downward.
  \item \textbf{faucet-close}: The robot arm has to use the red faucet handle and turn it anti-clockwise.
  \item \textbf{faucet-open}: The robot arm has to use the red faucet handle and turn it clockwise.
  \item \textbf{handle-press}: The robot arm has to press the red handle downward.
  \item \textbf{hammer}: The robot arm has to grip and pick up the hammer with a red handle and hit the peg on the box inside.
  \item \textbf{assembly}: The robot arm has to pick up the ring and place it into the red peg.
\end{itemize}
\end{promptbox}

\begin{figure*}[ht]
    \centering
    \includegraphics[width=0.95\linewidth]{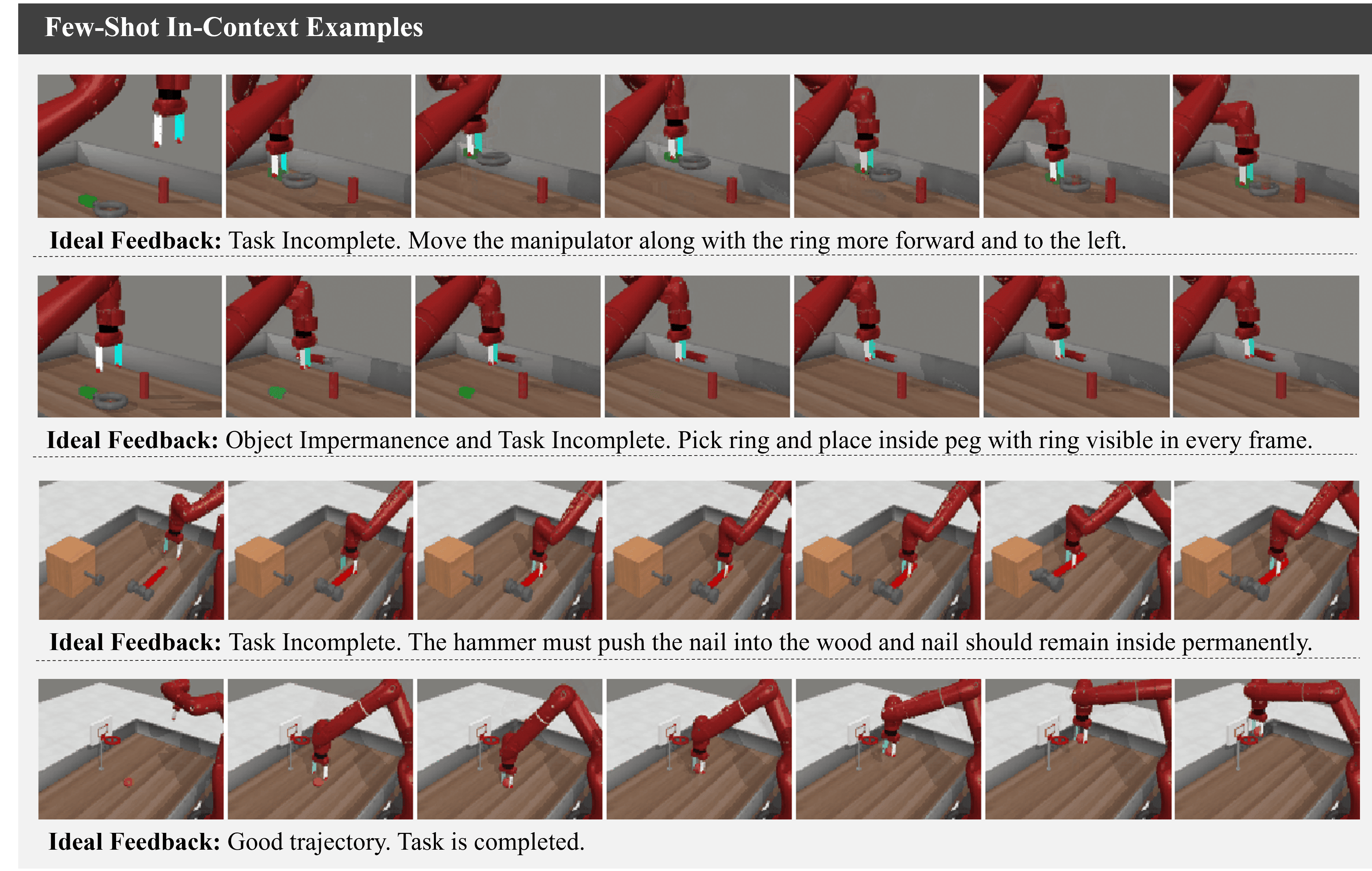}
    \caption{Few-Shot Examples given to VLM: We provide some examples to the VLM and corresponding feedback to teach the VLM in-context how to critic the generated videos for task completion and success or failure.}
    \label{fig:ICL}
\end{figure*}

\section{Dataset Descriptions in Detail}
\label{app:Data}

\begin{figure*}[ht]
    \centering
    \includegraphics[width=0.95\linewidth]{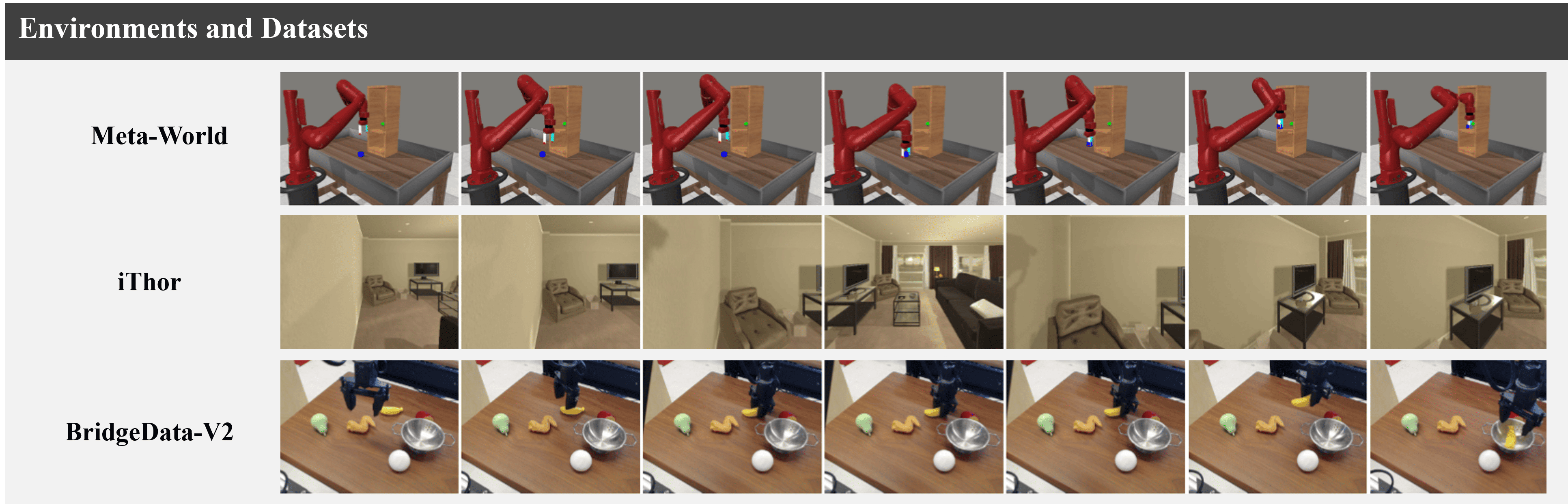}
    \caption{Environments and Datasets that we work with: Meta-World, iThor, and BridgeData-V2}
    \label{fig:datasets}
\end{figure*}

Meta-World ~\cite{yu2020meta} is a simulation benchmark that uses a Swayer robotic arm to perform a number of manipulation tasks. In our experiments, we make use of 11 tasks as shown in Table \ref{tab:mw_results}. We capture videos from three distinct camera angles for each task and use the same camera angles for both the training and testing phases. We gather five demonstration videos per task for each camera angle. During the evaluation, we tested on each of the three camera angles with 25 seeds per camera angle. The position of the robot arm and the object is randomized at the beginning of each seed to ensure variability. A trajectory is considered successful if the Video Agent reaches within a really close threshold of the goal state. 

iTHOR ~\cite{kolve2017ai2} is another popular 2D simulated benchmark that focuses on embodied common sense reasoning. We evaluate the Video as Agent framework on the object navigation tasks, where an agent is randomly initialized in a scene and tasked with finding an object of a specified type (e.g., toaster, television). At each time step, the agent can take one of the four possible actions (MoveForward, RotateLeft, RotateRight, or Done), and observes a 2D scene to operate in. We selected 12 objects ((e.g. toaster, television) to be placed in 4 different room types (e.g. kitchen, living room, bedroom, and bathroom). Again, the starting position of the agent is randomized at the start of each episode. During evaluation, we test the agent across 12 object navigation tasks spread across all 4 room types, 3 tasks per room. A trajectory is successful if the agent views and reaches within 1.5 meters of the target object before reaching the maximum environment step or predicting Done. 

To test the usefulness of our framework across different videos types, we also use the BridgeData V2 dataset ~\cite{walke2023bridgedata}, a large and diverse dataset of real world robotic manipulation behaviors designed to facilitate research in scalable robot learning. It contains 60,096 trajectories collected across 24 environments using a publicly available low-cost WidowX 250 6DOF robot arm. The dataset provides extensive task and environment variability, enabling skills learned from the data to generalize across environments and domains.

\subsection{Additional trajectories per iteration during online training}
We collect 15 successful trajectories for each task during every iteration. This standardization helps address task imbalance, as task success rates are higher for certain tasks compared to others. By ensuring a fixed number of successful trajectories per task, we prevent overfitting to easier tasks and maintain balanced model performance across the entire task set.

\section{Extended Experiments}\label{abl:extended-exp}
\subsection{Videos to action conversion}
We employ the GMFlow optical flow model to predict dense pixel movements across frames. These predicted flows serve as the foundation for reconstructing both object movements and robot motions depicted in the video. The flow predictions allow us to interpret the temporal evolution of the video in terms of actionable physical dynamics.
The optical flow essentially provides a dense correspondence of pixel movements between consecutive frames, which is then used to infer the relative motion of objects and the robot. This mapping bridges the gap between the high-dimensional video representation and the low-level control commands required to execute the tasks in a simulated or real environment.

This method ensures that the generated video plans are actionable and aligned with the task-specific dynamics, making the video generation process directly relevant to downstream policy learning and execution.

\subsection{Baseline experiments on Metaworld}\label{abl:baselines}
We conduct experiments on additional baselines including, Behavioral Cloning (BC), UniPi (with replan), VLP and Diffusion policy. Table \ref{tab:mw_baseline_results} consists of these results.

\begin{table*}[t]
\centering
\small
\setlength{\tabcolsep}{2pt}
\caption{\textbf{Meta-World Results.} The mean success rates of baselines and \method on 11 simulated robot manipulation environments from Meta-World. \method consistently outperforms baselines across all tasks.}
\scalebox{1}{
\begin{tabular}{l c c c c c c}
\toprule
& door-open & door-close & basketball & shelf-place & btn-press & btn-press-top \\
\midrule
BC-Scratch & 21.3\% & 36.0\% & 0.0\% & 0.0\% & 34.7\% & 12.0\% \\ 
BC-R3M & 1.3\% & 58.7\% & 0.0\% & 0.0\% & 36.0\% & 4.0\% \\ 
UniPi (with Replan) & 0.0\% & 36.0\% & 0.0\% & 0.0\% & 6.7\% & 0.0\% \\ 
AVDC & 30.7\% & 28.0\% & 21.3\% & 8.0\% & 34.7\% & 17.3\% \\ 
VLP & 33.3\% & 28.0\% & 17.3\% & 8.0\% & 36.0\% & 18.7\% \\ 
Diffusion Policy & 45.3\% & 45.3\% & 8.0\% & 0.0\% & 40.0\% & 18.7\% \\ 
AVDC-Replan & 72.0\% & 89.3\% & 37.3\% & 18.7\% & 60.0\% & 24.0\% \\ 
\midrule
VideoAgent & 40.0\% & 29.3\% & 13.3\% & 9.3\% & 38.7\% & 18.7\% \\ 
VideoAgent (Iter2) & 48.0\% & 40.0\% & 24.0\% & 12.0\% & 42.7\% & 36.0\% \\ 
VideoAgent (Iter3) & 58.7\% & 50.7\% & 28.0\% & 18.7\% & 53.3\% & 41.3\% \\ 
VideoAgent-Replan & \textbf{82.7\%} & \textbf{97.3\%}& \textbf{40.0\%} & \textbf{26.7\%} & \textbf{73.3\%} & \textbf{44.0\%}\\ 
\toprule
& faucet-close & faucet-open & handle-press & hammer & assembly & \cellcolor{gray!20}\textbf{Overall} \\
\midrule
BC-Scratch & 18.7\% & 22.7\% & 28.0\% & 0.0\% & 0.0\% & \cellcolor{gray!20}15.4\% \\ 
BC-R3M & 18.7\% & 17.3\% & 37.3\% & 0.0\% & 1.3\% & \cellcolor{gray!20}16.2\% \\ 
UniPi (with Replan) & 4.0\% & 9.3\% & 13.3\% & 4.0\% & 0.0\% & \cellcolor{gray!20}6.1\% \\ 
AVDC & 12.0\% & 17.3\% & 41.3\% & 0.0\% & 5.3\% & \cellcolor{gray!20}19.6\% \\ 
VLP & 30.7\% & 10.7\% & 33.3\% & 0.0\% & 1.3\% & \cellcolor{gray!20}19.8\% \\ 
Diffusion Policy & 22.7\% & \textbf{58.7\%} & 21.3\% & 4.0\% & 1.3\% & \cellcolor{gray!20}24.1\% \\ 
AVDC-Replan & 53.3\% & 24.0\% & 81.3\% & 8.0\% & 6.7\% & \cellcolor{gray!20}43.1\% \\ 
\midrule
VideoAgent & 46.7\% & 12.0\% & 36.0\% & 0.0\% & 1.3\% & \cellcolor{gray!20}22.3\% \\ 
VideoAgent (Iter2) & 53.3\% & 28.0\% & 52.0\% & 1.3\% & 5.3\% & \cellcolor{gray!20}31.2\% \\ 
VideoAgent (Iter3) & 58.7\% & 36.0\% & 64.0\% & 1.3\% & 9.3\% & \cellcolor{gray!20}38.2\% \\ 
VideoAgent-Replan & \textbf{74.7\%} & \textbf{46.7\%} & \textbf{86.7\%} & \textbf{8.0\%} & \textbf{10.7\%} & \cellcolor{gray!20}\textbf{53.7\%}\\ 

\bottomrule
\end{tabular}
}
\label{tab:mw_baseline_results}
\end{table*}

\subsection{Further Analysis of VideoAgent-Online}\label{abl:videoagent-online}
We train VideoAgent-Online for multiple iterations and observe that after 2 iterations, the results start to stabilize. The extra results for iteration 3 are also shown in table \ref{abltab:online}.

\begin{table*}[ht]
\centering
\small
\setlength{\tabcolsep}{3pt}
\scalebox{1}{

\begin{tabular}{l c c c c c c}
\toprule
& door-open & door-close & basketball & shelf-place & btn-press & btn-press-top \\
\midrule
AVDC & 30.7\% & 28.0\% & 21.3\% & 8.00\% & 34.7\% & 17.3\% \\ 
\midrule
VideoAgent & 40.0\% & 29.3\% & 13.3\% & 9.3\% & 38.7\% & 18.7\% \\ 
VideoAgent-Online (Iter1) & 48.0\% & 40.0\% & 24.0\% & 12.0\% & 42.7\% & 36.0\% \\ 
VideoAgent-Online (Iter2) & 58.7\% & 50.7\% & 28.0\% & 18.7\% & 53.3\% & 41.3\% \\ 
VideoAgent-Online (Iter3) & 58.7\% & 52.0\% & 26.7\% & 17.3\% & 54.7\% & 40.0\% \\ 
\midrule
\midrule

& faucet-close & faucet-open & handle-press & hammer & assembly & \cellcolor{gray!20}\textbf{Overall} \\
\midrule
AVDC & 12.0\% & 17.3\% & 41.3\% & 0.00\% & 5.30\% & \cellcolor{gray!20}19.6\% \\ 
\midrule
VideoAgent & 46.7\% & 12.0\% & 36.0\% & 0.0\% & 1.3\% & \cellcolor{gray!20}22.3\% \\ 
VideoAgent-Online (Iter1) & 53.3\% & 28.0\% & 52.0\% & 1.3\% & 5.3\% & \cellcolor{gray!20}31.2\% \\ 
VideoAgent-Online (Iter2) & 58.7\% & 36.0\% & 64.0\% & 1.3\% & 9.3\% & \cellcolor{gray!20}38.2\% \\ 
VideoAgent-Online (Iter3) & 56.3\% & 36.0\% & 66.7\% & 1.3\% & 10.7\% & \cellcolor{gray!20}38.22\% \\ 
\bottomrule
\end{tabular}
}
\caption{\textbf{Meta-World Result.} The mean success rates of VideoAgent combined with successive rounds of data collection via Online Iterations and Replan modules as compared to AVDC baseline.}

\label{abltab:online}
\end{table*}

\section{Architectural Details of VideoAgent}\label{abl:architecture}
\subsection{Video Diffusion training details}
We use the same video diffusion architecture as the AVDC baseline. For all models, we use dropout=0, num head channels=32, train/inference timesteps=100, training objective=predict v, beta schedule=cosine, loss function=l2, min snr gamma=5, learning rate=1e-4, ema update steps=10, ema decay=0.999. 

\subsection{Inference time speed}
In our current setup, during inference, our video generation model produces a new video within 10 seconds on a single A6000 GPU at a resolution of $128\times128$ for Meta-World. The process of mapping this generated video to an action takes, on average, an additional 25 seconds. This action-mapping stage involves calculating optical flow, receiving feedback from the vision-language model (VLM), and converting the video into an action sequence based on the computed flow.

\section{VLM Feedback for Correction}\label{abl:vlm_feedback}
\begin{table*}[ht]
\centering
\small
\setlength{\tabcolsep}{2pt}
\scalebox{1}{
\begin{tabular}{l c c c c c c}
\toprule
& door-open & door-close & basketball & shelf-place & btn-press & btn-press-top \\
\midrule
AVDC & 30.7\% & 28.0\% & 21.3\% & 8.00\% & 34.7\% & 17.3\% \\ 
\midrule
VideoAgent & 40.0\% & 29.3\% & 13.3\% & 9.3\% & 38.7\% & 18.7\% \\ 
VideoAgent-Binary & 46.7\% & 32.0\% & 14.7\% & 6.7\% & 38.7\% & 21.3\% \\ 
VideoAgent-Suggestive & 46.7\% & 33.3\% & 18.7\% & 12.0\% & 41.3\% & 22.7\% \\ 
VideoAgent-Online-Suggestive & 52.0\% & 28.0\% & 21.3\% & 16.0\% & 46.7\% & 22.7\% \\ 

\midrule
\midrule

& faucet-close & faucet-open & handle-press & hammer & assembly & \cellcolor{gray!20}\textbf{Overall} \\
\midrule
AVDC & 12.0\% & 17.3\% & 41.3\% & 0.00\% & 5.30\% & \cellcolor{gray!20}19.6\% \\ 
\midrule
VideoAgent & 46.7\% & 12.0\% & 36.0\% & 0.00\% & 1.3\% & \cellcolor{gray!20}22.3\% \\ 
VideoAgent-Binary & 46.7\% & 17.3\% & 32\% & 0.00\% & 5.3\% & \cellcolor{gray!20}23.8\% \\ 
VideoAgent-Suggestive & 48.7\% & 17.3\% & 46.7\% & 0.00\% & 5.3\% & \cellcolor{gray!20}26.6\% \\
VideoAgent-Online-Suggestive & 45.3\% & 20.0\% & 48.0\% & 2.7\% & 5.3\% & \cellcolor{gray!20}27.4\% \\

\bottomrule
\end{tabular}
}
\caption{\textbf{Meta-World: VideoAgent-Feedback Guided Results} The mean success rates for various tasks, comparing different VideoAgent-Feedback Guided variants and the AVDC baseline.}
\label{abltab:mw_feedback_resuts}
\end{table*}

\section{Details of Human Evaluation on BridgeData V2}
\label{app: human_eval_finegrained}
\paragraph{Qualitative Evaluation.} Next, we qualitatively evaluate video generation quality using the five Video-Score dimensions: Visual Quality (VQ) for clarity and resolution, Temporal Consistency (TC) for smooth frame transitions, Dynamic Degree (DD) for capturing accurate object/environment changes, Text-to-Video Alignment (TVA) for matching the video to the prompt, and Factual Consistency (FC) for adherence to physical laws and real-world facts. Videos are rated on a 4-point scale based on the metric in \cite{he2024videoscore}: 1 (Bad), 2 (Average), 3 (Good), and 4 (Perfect). Our evaluation is based on 50 generated videos from a held-out set. 

\begin{table*}[ht]
    \centering
    \caption{Task Success and Other Fine-grained Human Evaluation Metrics on BridgeData-V2}
    \begin{tabular}{l c c c}
        \toprule
        \textbf{Metrics} & &\textbf{AVDC} & \textbf{Video Agent} \\
        \midrule
        \rowcolor{gray!20} \textbf{Task Success via Human Eval} & & 42.0\%& \textbf{64.0\%} \\

        \midrule
        \multirow{5}{*}{\textbf{Holistic Assessment via Human Eval}} 
        & Visual Quality & 1.74 & 1.84\\
        & Temporal Consistency & 1.58 &1.76 \\
        & Dynamic Degree & 3.14 & 2.98 \\
        & Text to Video Alignment & 2.66 &  3.04 \\
        & Factual Consistency & 3.22 & 3.30  \\
        \midrule
        \rowcolor{gray!20} & \textbf{Human Eval Average} & 2.47 & \textbf{2.98} \\
        \bottomrule
    \end{tabular}
    \label{table:bridge_human_eval}
\end{table*}

In terms of VQ and TC, both the baseline AVDC and our VideoAgent generate average quality videos (graded 2), with AVDC hallucinating more and generating some choppy jumps in videos temporally (we grade such videos as 1) and Video Agent fixing some of these upon video conditioned iterative refinement.
The reason for AVDC baseline having higher DD is attributed to unruly movements that cause higher DD scores compared to VideoAgent, where movements are smoother. This also explains the result in fifth row of Table \ref{table:bridge_metrics_comparison}, and upon closer examination of the generated videos and their corresponding individual scores, we observed similar traits in videos having higher DD due to unnatural robot arm movements and object impermanence. TVA shows trends similar to ClipScore in Table \ref{table:bridge_metrics_comparison} due to the better instruction following ability of VideoAgent leading to more controlled generation. FC is a very crucial metric for deployment of video generation agents as policy for task completion in robotics, scene navigation, and so on. Improved visual quality does not imply adherence to correct physical laws and real-world constraints, FC particularly checks for this aspect and due to video conditioned self-refinement, VideoAgent has better FC compared to AVDC.

\clearpage
\newpage

\section{Examples}\label{abl:examples}
\subsection{Zero-shot generalization on real-world scenes}
\begin{figure*}[ht]
    \centering
    \includegraphics[width=0.95\linewidth]{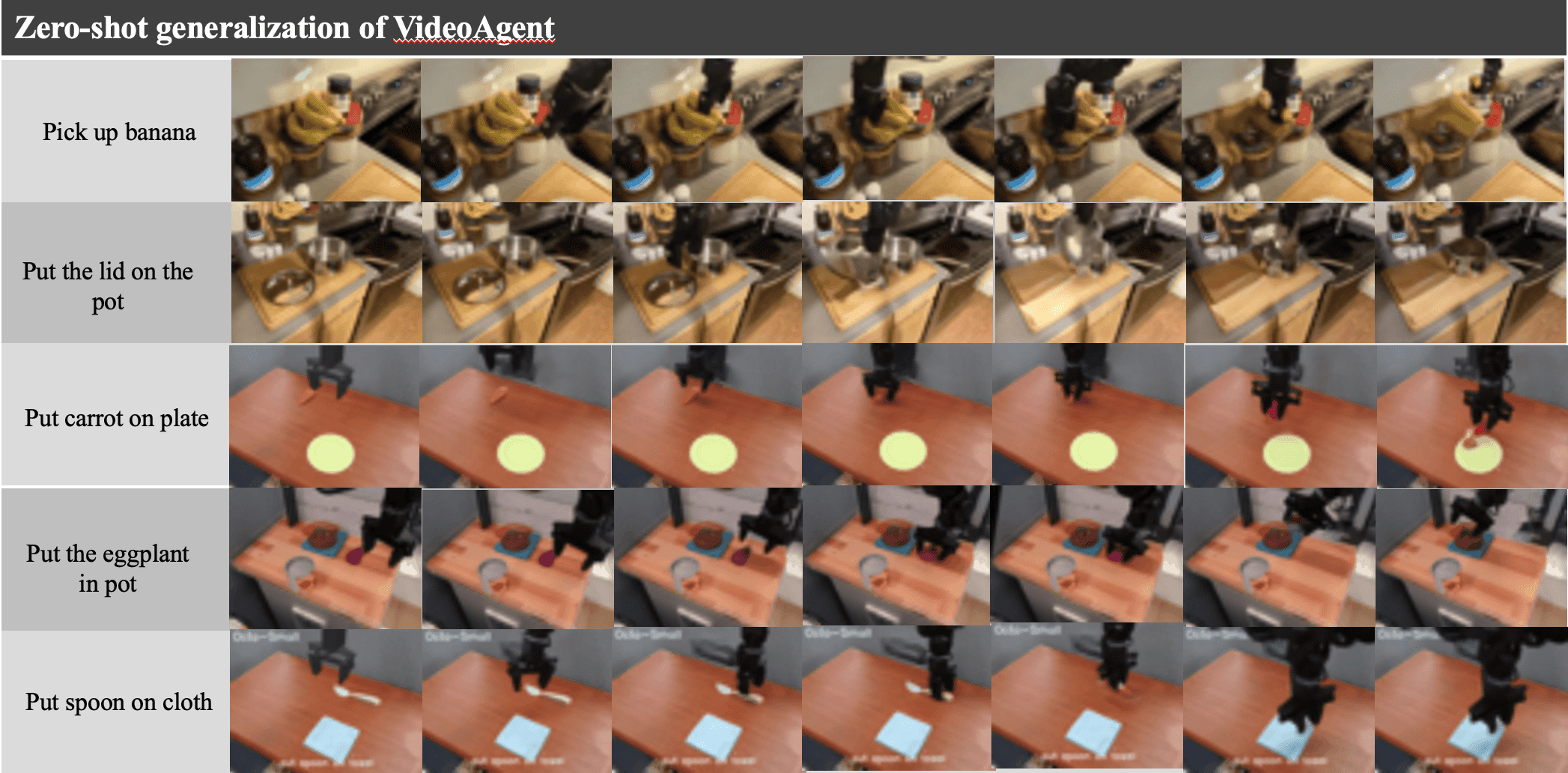}
    \caption{\textbf{Zero-shot generalization of VideoAgent:} VideoAgent generalizes fairly well to natural distribution shifts and is able to generate successful trajectories on data it has not been trained on.}
    \label{fig:bridge_generalization}
\end{figure*}

VideoAgent trained on Bridge dataset demonstrates strong performance on zero shot video generation for natural distribution shifts and longer language instructions. Some examples of the synthesized videos can be found in Fig. \ref{fig:bridge_generalization}.

\subsection{Improvements in Meta-World}
\begin{figure*}[ht]
    \centering
    \includegraphics[width=0.95\linewidth]{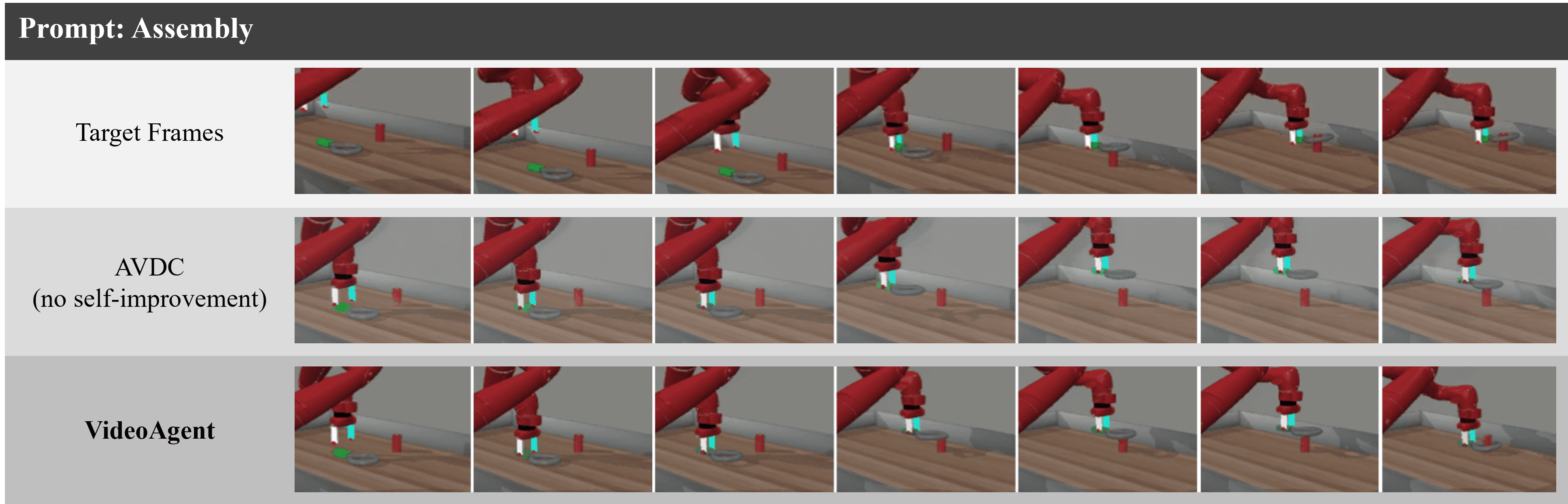}
    \caption{\textbf{Correcting Hallucinations in Video Generation:} The goal prompt is ``Assembly” as shown in the Target Video. The AVDC model has problem of object permanence and action incomplete in last frame. In contrast, our VideoAgent model accurately object permanence and correctly places the inside the peg properly.}
    \label{fig:metaworld_qual}
\end{figure*}
\subsection{Improvements in iThor}
\begin{figure*}[ht]
    \centering
    \includegraphics[width=0.95\linewidth]{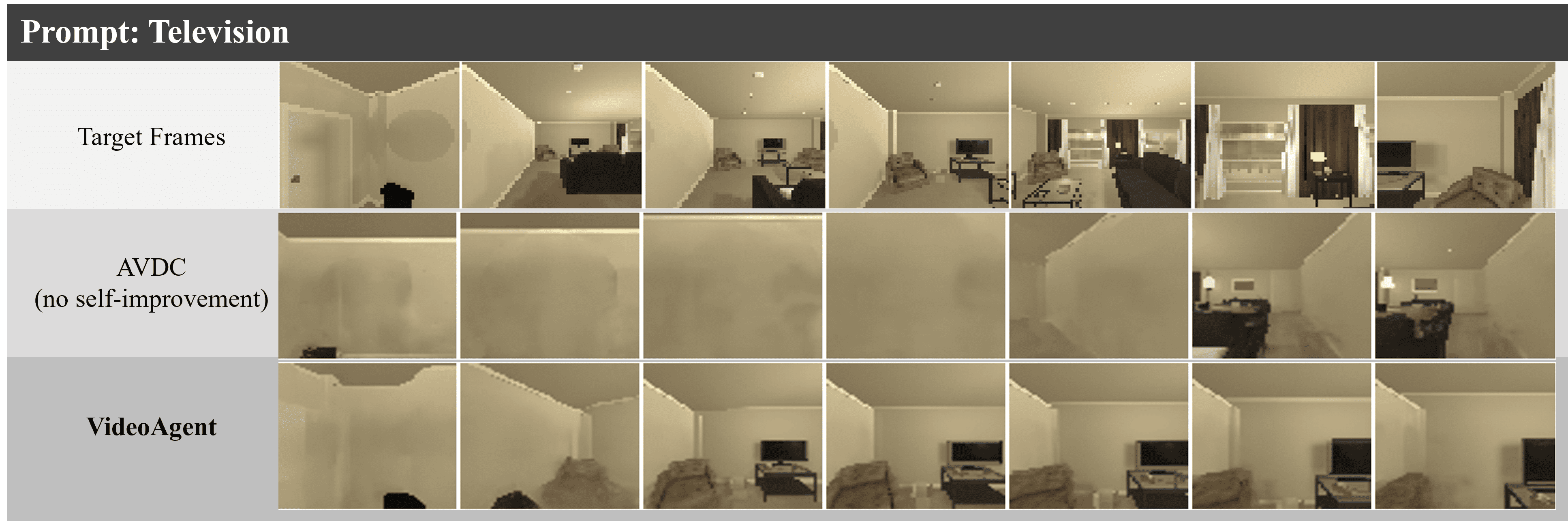}
    \caption{\textbf{Correcting Hallucinations in Video Generation:} The goal prompt is ``Television” as shown in the Target Video, the goal is for the navigator to locate the object and reach near it. The AVDC model has difficulty reconstructing and navigating in the livingroom to find the television. In contrast, our VideoAgent model solves the initial frame hallucinations and accurately reaches near the television correctly.}
    \label{fig:ithor_qual}
\end{figure*}
\clearpage
\subsection{Identification and Suggestive Feedback Examples}
\begin{figure*}[ht]
    \centering
    \includegraphics[width=0.95\linewidth]{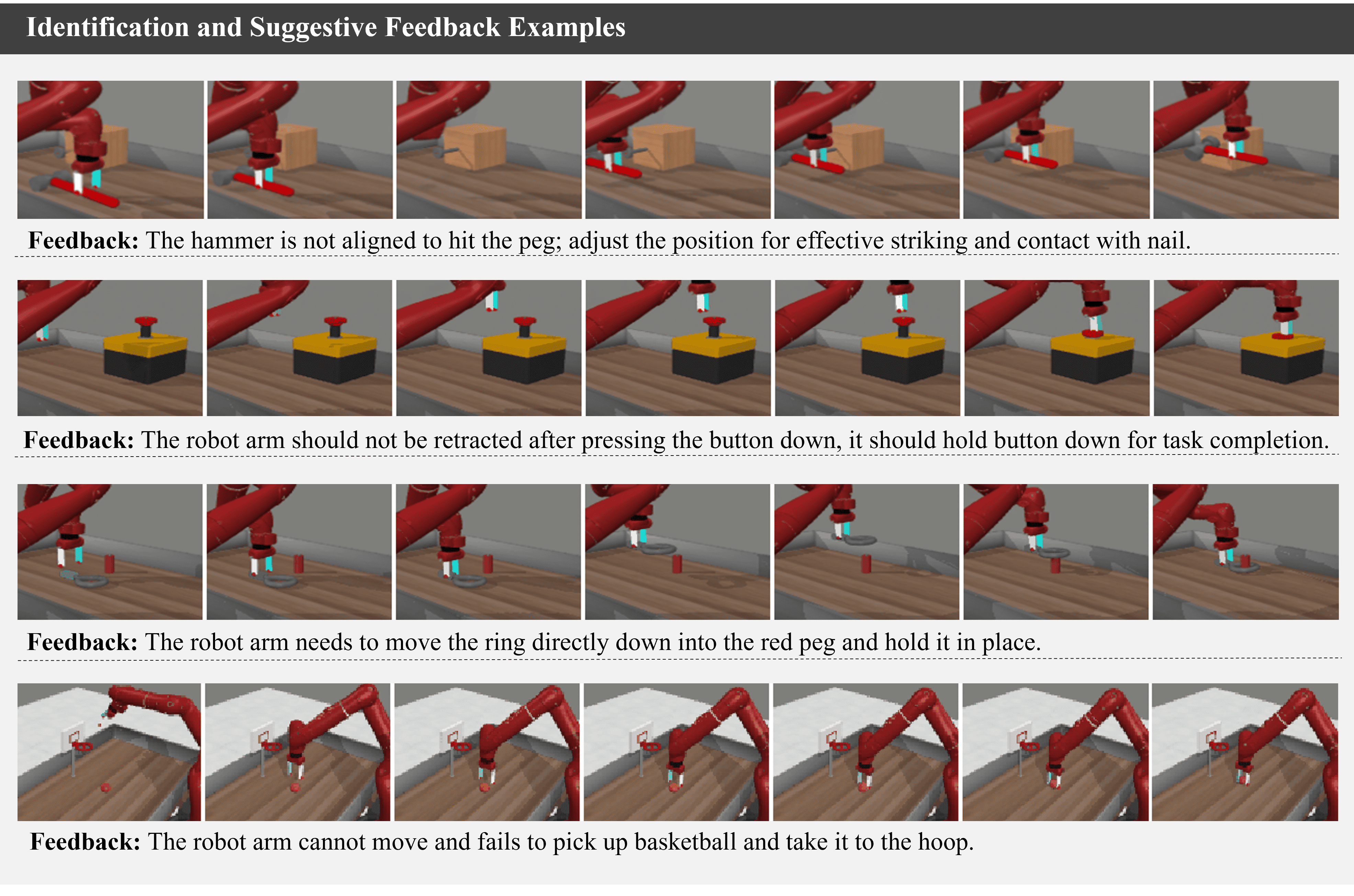}
    \caption{\textbf{Detailed VLM Feedback:} We show the efficacy of VLMs to provide useful feedback even in the absence of access to a simulator or real-world execution environment. The VLM acts as a proxy reward model to condition VideoAgent on useful corrective signals, leading to improved performance as described in Table \ref{tab:mw_feedback_results}.}
    \label{ablfig:suggestive_feedback}
\end{figure*}

\end{document}